\documentclass{article}
\usepackage[preprint,main]{spectranet_preprint}

\usepackage{amsmath,amssymb,amsthm}
\usepackage{graphicx}
\usepackage{booktabs}
\usepackage{multirow}
\usepackage[table]{xcolor}
\usepackage{hyperref}
\usepackage{cleveref}
\usepackage{microtype}
\usepackage{subcaption}
\usepackage{wrapfig}
\usepackage{enumitem}

\newtheorem{lemma}{Lemma}
\newtheorem{proposition}{Proposition}
\newtheorem{theorem}{Theorem}
\theoremstyle{remark}
\newtheorem{remark}{Remark}

\title{Bridging Spectral Operator Learning and U-Net Hierarchies:\\
SpectraNet for Stable Autoregressive PDE Surrogates}

\author{%
  Enrique Hern\'andez Noguera$^{1}$,\;
  Md Meftahul Ferdaus$^{1}$,\;
  Elias Ioup$^{2}$,\;
  Mahdi Abdelguerfi$^{1}$, \\
  Julian Simeonov$^{2}$ \\[10pt]
  \normalfont%
  $^{1}$University of New Orleans, New Orleans, LA, USA \\
  $^{2}$Naval Research Laboratory, Stennis Space Center, MS, USA \\[3pt]
  \texttt{\{ehernan8, mferdaus, gulfsceidirector\}@uno.edu}
}

\date{}

\begin{document}
\maketitle

\begin{abstract}
Neural operators for time-dependent PDEs face a structural tension: spectral architectures
(FNO and descendants) inherit exponential rollout-error growth from their one-step Lipschitz
constant, while hierarchical U-Net operators trade resolution invariance for multi-scale
detail. We introduce \emph{SpectraNet}, an autoregressive neural operator that composes
truncated spectral convolutions inside a U-Net hierarchy with a \emph{Residual-Target
Spectral Block} trained under a \emph{Semigroup-Consistency Loss}. The residual-target
parametrization replaces $L^T$ stability blow-up with linear $T\delta$ drift
(Theorem~\ref{thm:approx-stab}, conditional on staying in $\mathcal{K}$), and the
spectral path's parameter count is $\Theta(L w^2 M^2)$, independent of grid $N$.
Under a single unified protocol against $16$ published neural-operator baselines on
Navier--Stokes $\nu{=}10^{-5}$ at $64{\times}64$, SpectraNet reaches test relative
$L^2{=}0.0822$ at $2.04\,\text{M}$ parameters --- $2.33{\times}$ fewer than canonical
FNO at $\approx\!20\%$ lower error --- and wins five of six rows in a cross-PDE
comparison against FNO (NS at $\nu\!\in\!\{10^{-4},10^{-3}\}$, PDEBench Shallow-Water
2D and Diffusion-Reaction, with the Active-Matter row going to FNO inside its seed
spread). Trained from scratch at native $128^2$ under the same protocol, SpectraNet
improves to $0.0724$ while FNO regresses to $0.3080$; we caveat this in
\Cref{sec:results:resolution} since the unified protocol is not tuned per-architecture
per-resolution. Free rollout stays bounded for $T{=}100$ where FNO diverges across all
$200$ test trajectories. On consumer CPU at $B{=}1$, SpectraNet runs sub-$200\,\text{ms}$
while the full-attention Transformer that wins raw $L^2$ pays $\sim\!60{\times}$
latency; we do not claim to beat that Transformer on raw $L^2$, only to dominate the
lightweight ($\le\!5\,\text{M}$ parameter, sub-$200\,\text{ms}$ CPU) Pareto frontier. Source code is available at \url{https://github.com/Enrikkk/spectranet}.
\end{abstract}

\section{Introduction}\label{sec:intro}

Neural operators amortize the cost of solving time-dependent partial differential
equations (PDEs)~\citep{li2020fno,kovachki2023neural,schmidt2026algorithm} by learning the solution operator
$\Phi_{\Delta t}: \omega_t \!\mapsto\! \omega_{t+\Delta t}$ from data, then rolling it
out autoregressively (AR) to predict trajectories. Two failure modes dominate this
regime. First, any one-step error compounds across $T$ rollouts at a rate set by the
operator's one-step Lipschitz constant; for direct-prediction architectures this is
generically exponential in $T$. Second, existing operators choose between resolution
invariance and multi-scale detail rather than combining them: spectral architectures
(the Fourier Neural Operator, FNO~\citep{li2020fno}, and its descendants) inherit a
fixed band-limit but no spatial hierarchy; U-Net~\citep{ronneberger2015unet} hybrids
recover detail but lose resolution invariance at down/up-sampling; and transformer
operators replace the spectral mixer with attention that scales as $\mathcal{O}(N^2)$
in the spatial token count $N$.

\paragraph{Gap.} No published neural operator combines all four ingredients that
\Cref{thm:approx-stab} and \Cref{prop:param-eff} below identify as jointly necessary at
the $64^2$--$256^2$ regime: \textbf{(i)} truncated spectral mixing for resolution-aware
approximation, \textbf{(ii)} a U-Net hierarchy for multi-scale capture,
\textbf{(iii)} a residual-target rollout parametrization that bounds long-horizon drift,
and \textbf{(iv)} a training objective that enforces semigroup consistency at the
trajectory level. FNO has only (i); U-FNO~\citep{wen2022ufno} and U-NO~\citep{rahman2023uno}
have (i)+(ii); transformers have neither; (iii) and (iv) have not been combined with
spectral or U-Net mixers in the AR-rollout regime.

\paragraph{SpectraNet.} We introduce \emph{SpectraNet}, an autoregressive neural operator
that bridges spectral operator learning and U-Net hierarchies through a
\emph{Residual-Target Spectral Block} that parametrizes the per-step update as
$f_\theta = \mathrm{id} + \Delta_\theta$, together with a
\emph{Semigroup-Consistency Loss} that ties two single-step applications
\mbox{$f_\theta\!\circ\!f_\theta(\omega_t)$} to the two-step ground truth $\omega_{t+2}$. The headline operating point uses three
encoder--decoder levels, channel width $w{=}32$, spectral truncation radius $M{=}12$
(the number of low-frequency Fourier modes retained per axis), and $2.04\,\text{M}$
parameters --- $2.33{\times}$ fewer than canonical FNO at $\approx\!20\%$ lower error
on Navier--Stokes (NS) at viscosity $\nu = 10^{-5}$.

\paragraph{Contributions.}
\textbf{(C1) Architecture (\Cref{sec:method}):} a 2D autoregressive spectral U-Net with
the two named training-time innovations above; the canonical operating point sits on the
lightweight ($\le\!5\,\text{M}$ parameter, sub-$200\,\text{ms}$ CPU) Pareto frontier of
the NS $\nu{=}10^{-5}$ leaderboard, dominating every published baseline within that
envelope (\Cref{tab:leaderboard}). The full-attention NSL Transformer wins raw $L^2$
outside this envelope and we do not claim to beat it on accuracy.
\textbf{(C2) Theory (\Cref{sec:theory}):} an Approximation--Stability decomposition
(\Cref{thm:approx-stab}) bounding the $T$-step stability term by $\mathcal{O}(T \delta)$
under residual-target parametrization vs $\mathcal{O}(L^T \varepsilon_0)$ for direct
prediction with one-step Lipschitz $L\!>\!1$; \Cref{lemma:residual-stab}(b) is the standard
residual-network observation, and the contribution is anchoring it to an empirical operator
with measured $\delta, \widehat{L}$ (\Cref{tab:lipschitz}). \Cref{prop:param-eff} bounds
the spectral path at $\Theta(L w^2 M^2)$, independent of $N$ (shared with FNO).
\textbf{(C3) Multi-PDE evidence (\Cref{sec:results:crosspde}):} a unified-protocol
evaluation over six PDE benchmarks in which SpectraNet beats FNO on five of six rows at
$2.33{\times}$ smaller parameter count; the NS row is two-seed and the remaining five
are single-seed (\Cref{sec:discussion:limitations}). At native $128^2$ under the same
protocol SpectraNet improves to $0.0724$ while FNO regresses to $0.3080$; caveat in
\Cref{sec:results:resolution}.
\textbf{(C4) Deployment cost (\Cref{sec:timing}):} a CPU-extended inference-cost study
showing SpectraNet runs sub-$200\,\text{ms}$ on consumer CPU while the full-attention
Transformer that wins raw $L^2$ takes $\sim\!10\,\text{s}$ per sample.

\section{Related Work}\label{sec:related}

We position SpectraNet against four prior families of regular-grid neural operators,
each embodying one or two of the four ingredients from \Cref{sec:intro}.

\textbf{Spectral neural operators.} The Fourier Neural Operator~\citep{li2020fno} is the
ancestor of SpectraNet's spectral path: pointwise channel mixes interleaved with
truncated spectral convolutions, granting (i) resolution invariance and translation
equivariance up to the bandlimit \citep{ferdaus2025comprehensive}. Variants address specific failure modes:
F-FNO~\citep{tran2021ffno} factorizes 2D spectral convolutions for memory;
U-FNO~\citep{wen2022ufno} adds a U-Net hierarchy on top; U-NO~\citep{rahman2023uno}
extends with multi-resolution spectral mixing; LSM~\citep{wu2023lsm} learns a basis;
MWT~\citep{gupta2021mwt} uses multi-wavelets. SpectraNet inherits the FNO spectral
convolution but adds (a) the Residual-Target Spectral Block (\Cref{sec:method:residual})
and (b) the Semigroup-Consistency Loss (\Cref{sec:method:twostep}). U-FNO/U-NO add the
hierarchy but neither has a residual-target rollout or a multi-step training objective;
both inherit the direct-prediction $L^T$ stability blow-up of \Cref{thm:approx-stab}(i).

\textbf{Transformer operators.} Galerkin~\citep{cao2021galerkin}, FactFormer~\citep{li2023factformer},
OFormer~\citep{li2023oformer}, Transolver~\citep{wu2024transolver}, GNOT~\citep{hao2023gnot}
all replace the spectral mixer with sub-$\mathcal{O}(N^2)$ approximate attention. Only
the NSL Transformer~\citep{wu2024nsl} retains exact softmax and clears FNO at $64^2$ ---
a resolution-dependent result documented in \Cref{appendix:attention-discussion}.
\Cref{prop:param-eff} predicts these transformers' parameter cost grows polynomially with
grid resolution, borne out by their $39$--$277\,\text{M}$ counts (\Cref{tab:leaderboard}).

\textbf{Other regular-grid operators.} CNO~\citep{raonic2023cno}, KNO~\citep{xiong2024kno},
and ONO~\citep{xiao2024ono} round out the class. We exclude graph operators (GINO, MP-PDE)
and foundation models (Poseidon, DPOT) as out-of-scope.

\textbf{Stability of autoregressive operators.} Long-horizon stability has been analyzed
via spectral norms~\citep{kovachki2023neural}, dissipation matching~\citep{lippe2023pderefiner},
and physics-informed regularizers~\citep{wang2021pinn}. \citet{lippe2023pderefiner} relate
rollout instability to the operator's one-step Lipschitz constant; we adapt this view in
\Cref{lemma:residual-stab} to contrast direct vs residual-target prediction.

\textbf{Residual-target rollout and multi-step training.} Residual-target prediction
($f \!=\! \mathrm{id} \!+\! \Delta_\theta$) is a classical identity prior~\citep{he2016resnet}
and has been used in autoregressive PDE surrogates before:
\citet{brandstetter2022mppde} adopt the difference parametrization in MP-PDE on graphs,
and \citet{lippe2023pderefiner} place a denoising correction on top of one-step
prediction. The linear-drift observation in \Cref{lemma:residual-stab}(b) is the
standard residual-network fact and is not new on its own \cite{pokhrel2020forecasting}. Multi-step / pushforward
training objectives that tie composed predictions to multi-step ground truth also have
prior instances --- the pushforward trick in \citet{brandstetter2022mppde} and the
iterative refinement of \citet{lippe2023pderefiner} are close cousins of our two-step
Semigroup-Consistency Loss. Our contribution is the specific combination: residual-target
rollout embedded in a truncated-spectral U-Net, trained with the two-step semigroup
penalty, evaluated under a unified protocol against $17$ baselines, with $\delta$ and
$\widehat{L}$ measured on the same benchmark
(\Cref{tab:lipschitz,sec:results:longhorizon}) against a direct-prediction baseline
that catastrophically diverges at the same resolution.

\section{Problem Formulation}\label{sec:setup}

Let $\Phi_{\Delta t}: H^s(\Omega) \!\to\! H^s(\Omega)$ be the time-$\Delta t$ flow of a
PDE on a spatial domain $\Omega$. We seek a parametric \emph{neural operator}
$f_\theta: \mathbb{R}^{H \times W \times T_{\text{in}}} \!\to\! \mathbb{R}^{H \times W}$
that approximates the one-step flow on a discretization: given a sliding input window
of $T_{\text{in}}{=}10$ past frames, $f_\theta$ predicts the next frame. At test time
$f_\theta$ is applied autoregressively to produce a free rollout of horizon
$T_{\text{out}}{=}10$ frames.
Training minimizes the per-sample relative $L^2$ loss
$\mathcal{L}_{L^2}(\widehat{\omega}, \omega) \!=\! \|\widehat{\omega} - \omega\|_2 / \|\omega\|_2$
with single-step teacher forcing; the headline metric is the joint trajectory relative
$L^2$ over all $T_{\text{out}}$ test frames stacked.

\paragraph{Rollout-error decomposition.} Let
$\mathcal{E}(T) \!=\! \mathbb{E}_{\omega_0}[\|f_\theta^T(\omega_0) - \Phi^T(\omega_0)\|]$
be the expected $T$-step rollout error. Standard
arguments~\citep{kovachki2023neural,lippe2023pderefiner} decompose
\begin{equation}
\mathcal{E}(T) \;=\; \underbrace{\mathcal{E}_{\mathrm{approx}}(T)}_{\text{representation gap}}
                   \,+\, \underbrace{\mathcal{E}_{\mathrm{stab}}(T)}_{\text{rollout amplification}}\,.
\label{eq:rollout-decomp}
\end{equation}
\Cref{thm:approx-stab} formalizes how SpectraNet controls both terms simultaneously.

\paragraph{Desiderata.}\label{sec:setup:desiderata} A regular-grid neural operator that aims to be Pareto-dominant
across cost, accuracy, and rollout horizon should satisfy: \textbf{(D1)} resolution-aware
approximation --- $\mathcal{E}_{\mathrm{approx}}$ depends on smoothness and truncation
budget but not on grid $N$; \textbf{(D2)} bounded long-horizon drift ---
$\mathcal{E}_{\mathrm{stab}}(T)$ grows at most linearly in $T$, ruling out architectures
with $L^T$ blow-up; \textbf{(D3)} parameter efficiency --- the spatial mixer's parameter
count is bounded as $N \!\to\! \infty$; \textbf{(D4)} multi-scale detail capture beyond
a flat band-limit $M$. \Cref{sec:method} shows that combining a U-Net hierarchy (D4)
with truncated spectral convolutions (D1, D3) and a Residual-Target Spectral Block (D2)
satisfies all four, and \Cref{sec:theory} proves D2 (\Cref{thm:approx-stab}) and D3
(\Cref{prop:param-eff}).

\section{The SpectraNet Architecture}\label{sec:method}

\begin{figure}[ht]
\centering
\includegraphics[width=0.95\textwidth]{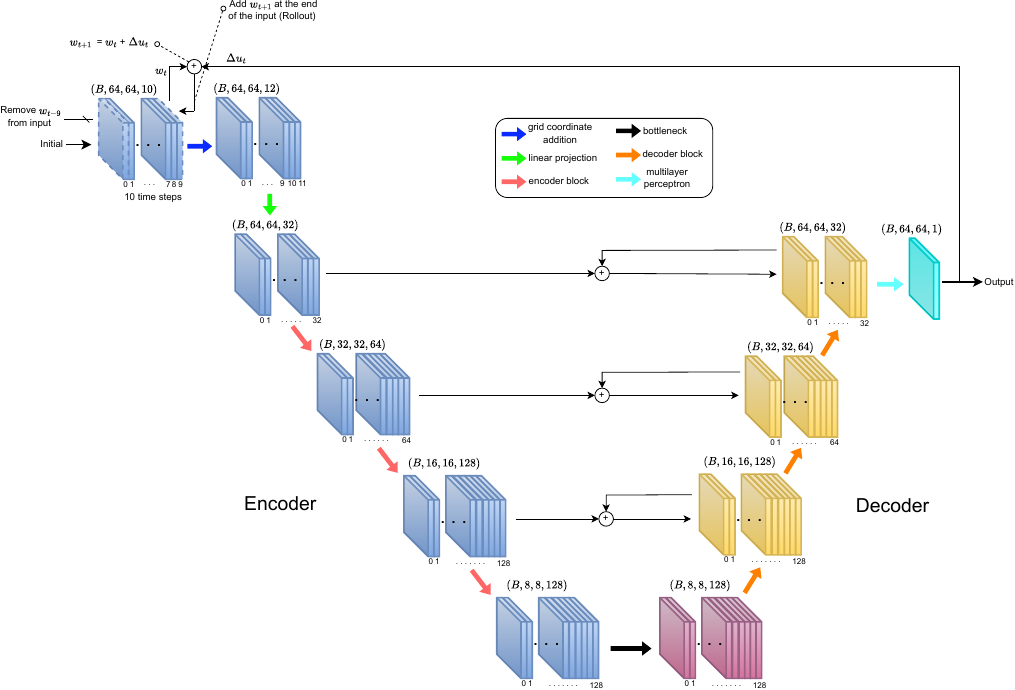}
\caption{\textbf{SpectraNet architecture.} Input $(B,64,64,10)$ + $(x,y)$-grid is
lifted to $w{=}32$ channels; a three-level encoder at truncations
$M_\ell\!\in\!\{12,6,3\}$ feeds a single-block bottleneck ($M_3{=}1$); a mirrored
decoder upsamples with additive skips ($\oplus$); a two-layer $1{\times}1$ MLP head
(hidden $4w$, GeLU) emits residual $\Delta_\theta$, summed with $\omega_t$ to recover
$\widehat{\omega}_{t+1}$ (\Cref{thm:approx-stab}).}
\label{fig:architecture}
\end{figure}

\subsection{Overview and named modules}\label{sec:method:arch}

\textbf{Base operator.} SpectraNet is a parametric map
$f_\theta : \mathbb{R}^{H \times W \times T_{\text{in}}} \!\to\! \mathbb{R}^{H \times W}$
that maps a $T_{\text{in}}{=}10$-frame window of vorticity to the next frame; the full
$T_{\text{out}}{=}10$ trajectory is produced by autoregressive rollout. The network combines
a spectral path (truncated Fourier convolutions) with a U-Net encoder--decoder backbone
(additive skips) and a residual-target output layer; layer-by-layer pseudocode and
parameter counts are in \Cref{appendix:arch}. SpectraNet introduces two named building
blocks: the \emph{Residual-Target Spectral Block} (\Cref{sec:method:residual}) and the
\emph{Semigroup-Consistency Loss} (\Cref{sec:method:twostep}); the output of the final
decoder level is mapped to the predicted vorticity by a two-layer $1{\times}1$ MLP head
(\Cref{sec:method:head}).

\subsection{Encoder--bottleneck--decoder backbone}\label{sec:method:backbone}

\textbf{Spectral block.} Each level $\ell$ takes a feature map
$x_\ell \in \mathbb{R}^{H_\ell \times W_\ell \times C_\ell}$ and applies a parallel
spectral path and channel-mixing path:
\begin{equation}
y_\ell \;=\; \mathrm{GeLU}\!\Big(\mathrm{Spec}_{M_\ell}(x_\ell) \,+\, \mathrm{MLP}_{1{\times}1}(x_\ell) \,+\, \mathrm{Conv}_{1{\times}1}(x_\ell)\Big),
\label{eq:enc-block}
\end{equation}
where $\mathrm{Spec}_{M_\ell}$ is a truncated spectral convolution at radius $M_\ell$,
implemented via the Fast Fourier Transform (FFT), with $M_\ell$ Fourier modes retained
per axis (\Cref{appendix:arch:spectral}); $\mathrm{MLP}_{1{\times}1}$ is a two-layer
pointwise multi-layer perceptron (MLP) with the GeLU non-linearity; and
$\mathrm{Conv}_{1{\times}1}$ is a residual identity branch. Encoder steps follow
each block by $2{\times}$ average-pool down-sampling and a $1{\times}1$ projection that
doubles the channel count up to a cap $4w$. Spectral truncation halves at each level,
$M_\ell \!=\! \min(\lfloor M / 2^\ell \rfloor,\, H_\ell/2)$, capped at the level Nyquist;
at the headline ($w{=}32$, $M{=}12$, $L{=}3$) this gives $M_\ell \!\in\! \{12, 6, 3, 1\}$
and the encoder/decoder shapes shown in \Cref{fig:architecture}. The bottleneck applies
\Cref{eq:enc-block} once at $(8, 8, 4w)$ with no down/up-sampling and no attention.
Each decoder step bilinearly upsamples, projects channels with a $1{\times}1$ conv,
merges the encoder skip additively (\emph{$\oplus$}), and applies \Cref{eq:enc-block}.

\subsection{Output projection}\label{sec:method:head}

\textbf{Two-layer $1{\times}1$ MLP head.} The final decoder feature map
$z \!\in\! \mathbb{R}^{64 \times 64 \times w}$ is mapped to the raw prediction
by $\widehat{\omega}_{t+1}^{\,\text{raw}} \!=\! W_2\,\mathrm{GeLU}(W_1\,z)$
with $W_1 \!\in\! \mathbb{R}^{4w \times w}$, $W_2 \!\in\! \mathbb{R}^{1 \times 4w}$
(\Cref{appendix:arch}); at $w{=}32$ this is $\!\approx\! 4.2\,\text{K}$
parameters ($\!\approx\! 0.2\%$ of the total). Two decorated alternatives ---
a multi-resolution head combining three per-level predictions via a learned softmax
weighting, and an efficient KAN layer in place of $W_2$ --- are tested in the output-head
ablation (\Cref{sec:ablations:micro}); neither improves accuracy and both are excluded
to keep the parameter count and the named-component list minimal.

\subsection{Residual-Target Spectral Block}\label{sec:method:residual}

\textbf{The central design choice.} SpectraNet does not predict $\omega_{t+1}$ directly:
the network's raw output is the per-step \emph{residual}
$\Delta_t \!=\! \omega_{t+1} \!-\! \omega_t$, and the integrated prediction is
\begin{equation}
\widehat{\omega}_{t+1} \;=\; \omega_t \,+\, f_\theta\big(\omega_{t-T_{\text{in}}+1{:}t}\big),
\qquad f_\theta = \mathrm{id} + \Delta_\theta.
\label{eq:residual}
\end{equation}
Training loss is per-sample relative $L^2$ on the raw residual, not on the integrated
prediction. This is the analogue of ResNet's identity-preconditioning~\citep{he2016resnet}
for spectral PDE operators. \Cref{thm:approx-stab}(i) shows it replaces the $L^T$
stability blow-up of direct-prediction operators with a linear-in-$T$ drift.
At test time we use free rollout (no teacher forcing); we do not employ scheduled sampling.

\subsection{Semigroup-Consistency Loss}\label{sec:method:twostep}

\textbf{Trajectory-level penalty on the operator semigroup.} In addition to the per-step
residual loss, we add a $\lambda$-weighted penalty tying two single-step applications to
the two-step ground truth:
\begin{equation}
\mathcal{L}_{\text{train}}
\,=\, \mathcal{L}_{L^2}(\widehat{\omega}_{t+1}^{\,\text{raw}}, \Delta_t)
\,+\, \lambda \, \mathcal{L}_{L^2}\big(f_\theta \!\circ\! f_\theta(\omega_{t-T_{\text{in}}+1{:}t}), \omega_{t+2}\big),
\label{eq:loss}
\end{equation}
with $\lambda \!=\! 0.1$. The penalty enforces the discrete operator-semigroup identity
$\Phi_{2 \Delta t} \!=\! \Phi_{\Delta t} \!\circ\! \Phi_{\Delta t}$ at training time. There is
no inference-time penalty: rollout is the same single-step AR loop. Empirically
(\Cref{sec:ablations}) this term contributes $-0.0029$ in trajectory $L^2$ at no
inference cost; \Cref{rmk:semigroup-lipschitz} explains why the improvement is invisible
to the empirical Lipschitz constant.

\section{Stability and Resolution Invariance}\label{sec:theory}

We prove two structural properties that distinguish SpectraNet from prior families.
\Cref{thm:approx-stab} bounds rollout error termwise via \Cref{eq:rollout-decomp};
\Cref{prop:param-eff} bounds parameter cost independently of grid. The component
bounds (\Cref{lemma:residual-stab,prop:res-inv}) and all proofs are in
\Cref{appendix:proofs}.

\begin{theorem}[Approximation--Stability decomposition]\label{thm:approx-stab}
Let $f_\theta$ approximate the time-$\Delta t$ flow $\Phi$ of an autonomous PDE on a
compact set $\mathcal{K} \!\subset\! H^s(\mathbb{T}^2)$. Then SpectraNet's design choices
control the two terms in \Cref{eq:rollout-decomp} separately:
\textbf{(i)} for direct prediction with one-step Lipschitz constant $L$,
$\mathcal{E}_{\mathrm{stab}}(T) \!=\! \mathcal{O}(L^T \varepsilon_0)$;
for SpectraNet's residual-target parametrization $f_\theta \!=\! \mathrm{id} + \Delta_\theta$
with $\delta \!=\! \sup_\mathcal{K} \|\Delta_\theta\|$,
$\mathcal{E}_{\mathrm{stab}}(T) \!=\! \mathcal{O}(T \delta)$ (\Cref{lemma:residual-stab}).
\textbf{(ii)} Under spectral truncation at radius $M$, $\mathcal{E}_{\mathrm{approx}}$
is bounded by the band-limited tail $\|\omega - P_M \omega\|_{H^s}$, independent of grid
$N \!\geq\! 2M$ (\Cref{prop:res-inv}). Consequently,
\begin{equation}
\mathcal{E}(T) \;\le\; C(s, M)\,\|\omega - P_M \omega\|_{H^s} \,+\, T \delta,
\label{eq:thm-bound}
\end{equation}
replacing the $L^T$ blow-up of direct-prediction operators with linear drift.
\end{theorem}

\begin{lemma}[Linear vs exponential rollout drift]\label{lemma:residual-stab}
Let $f : \mathbb{R}^N \!\to\! \mathbb{R}^N$ approximate $\Phi$ on $\mathcal{K}$.
\textbf{(a)} If $\sup \|f(u) - \Phi(u)\| \!\leq\! \varepsilon_0$ and $f$ is $L$-Lipschitz,
then for $L \neq 1$ the $T$-step error is bounded by
$\|f^T(u_0) - \Phi^T(u_0)\| \!\leq\! \varepsilon_0 (L^T \!-\! 1)/(L \!-\! 1)$, exponential
for $L \!>\! 1$.
\textbf{(b)} If $f(u) = u + \Delta(u)$ with $\sup \|\Delta(u)\| \leq \delta$, then
$\|f^T(u_0) - u_0\| \leq T \delta$.
\end{lemma}

\begin{proposition}[Resolution invariance of the spectral mixer]\label{prop:res-inv}
The spectral conv layer $f^{\mathrm{spec}}_M$ at truncation $M$ acts identically on
$\mathrm{rfft}$ coefficients inside the disk $\|k\| \!\leq\! M$ at any two grid resolutions
$N_1, N_2 \!\geq\! 2M$ of the same continuous initial condition. The band-limited contribution
to the layer's output is resolution-invariant up to bandlimit-aliasing residual
$\mathcal{O}(N_{\min}^{-1})$.
\end{proposition}

\begin{proposition}[Parameter efficiency of SpectraNet's spectral path]\label{prop:param-eff}
At width $w$, depth $L$, truncation $M$, grid $N$:
$P_{\mathrm{SpectraNet}}(w, L, M, N) \!=\! \Theta(L \cdot w^2 \cdot M^2)$, independent of
$N$. Full self-attention has parameter count $\Theta(L w^2)$ in projections plus
$\Theta(N^2)$ memory and compute per layer for the attention matrix itself; standard
convolution at kernel $k$ matching the spectral mixer's global reach requires
$L \!\sim\! N/k$ stacked layers, yielding $\Theta(w^2 N k)$ parameters. SpectraNet
shares the $N$-independent parameter count with the FNO family --- both are spectral
operators with the same Fourier-mode budget --- and inherits the $N$-independence
from this lineage; the proposition isolates this shared property in contrast to
attention and convolution-only mixers, not as a property unique to SpectraNet.
\end{proposition}

\textbf{Empirical anchor.} The empirical Lipschitz constant $\widehat{L}$ estimated via
Gaussian perturbations of size $10^{-3}$ is reported in \Cref{tab:lipschitz}
(\Cref{appendix:detailed-results}). FNO has $\widehat{L}(T{=}1)\!\approx\!1.84$;
\Cref{lemma:residual-stab}(a) is a worst-case bound predicting $L^T$-style amplification,
qualitatively consistent with the catastrophic divergence observed in
\Cref{sec:results:longhorizon} (we do not claim the worst-case bound is tight).
SpectraNet has $\delta \!\approx\! 0.026\|\omega\|$ measured along training trajectories;
(b) gives cumulative drift $\leq T \delta$ \emph{if} $\|\Delta_\theta\|$ remains bounded
along the rollout, with observed $\approx\!4.8{\times}$ energy growth at $T{=}100$.
SpectraNet's own $\widehat{L}\!\in\![8,11]$ in this probe is larger than FNO's; most of
that is the additive identity in $f \!=\! \mathrm{id}+\Delta_\theta$, which inflates the
local gradient probe near typical inputs.

\begin{remark}\label{rmk:semigroup-lipschitz}
\Cref{lemma:residual-stab}(b) bounds $\mathcal{E}_{\mathrm{stab}}$ via
$\delta \!=\! \sup_\mathcal{K} \|\Delta_\theta\|$ \emph{conditional on the rollout
trajectory remaining in the compact set $\mathcal{K}$ on which $\delta$ is taken}.
We do not prove this in-distribution property; we verify it empirically by running
free rollout to $T{=}100$ ($10{\times}$ the training horizon) and confirming SpectraNet's
energy remains bounded across all $200$ test trajectories
(\Cref{sec:results:longhorizon}). The Semigroup-Consistency Loss penalizes a
trajectory-level deviation, not $\sup\|\nabla f_\theta\|$; we observe $\widehat{L}$
unchanged by this loss (\Cref{tab:lipschitz}). More generally, for residual-target
operators $f \!=\! \mathrm{id}+\Delta_\theta$ a local-gradient probe like $\widehat{L}$
is inflated by the identity branch and is not the relevant stability quantity; the
relevant quantity is the trajectory-uniform residual bound $\delta$. We report
$\widehat{L}$ in \Cref{tab:lipschitz} as the load-bearing diagnostic for the
direct-prediction baselines (FNO and the NSL Transformer) and as a sanity diagnostic
for SpectraNet.
\end{remark}

\section{Experimental Setup}\label{sec:experiments}

\textbf{Datasets.} Headline benchmark: Navier--Stokes (NS) vorticity at viscosity
$\nu \!=\! 10^{-5}$, the $64{\times}64$ dataset
\texttt{NavierStokes\_V1e-5\_N1200\_T20.mat}~\citep{li2020fno} ($1200$ trajectories,
$20$ time steps; the operator-learning task is to predict an output horizon of
$T_{\text{out}}{=}10$ frames from an input window of $T_{\text{in}}{=}10$ past frames).
Cross-PDE evaluation uses NS at $\nu{=}10^{-3}, 10^{-4}$~\citep{li2020fno},
PDEBench Shallow-Water 2D and Diffusion-Reaction~\citep{takamoto2022pdebench}, and
The~Well Active-Matter~\citep{ohana2025well}. The native-$128^2$ resolution experiment
uses a $128{\times}128$ NS $\nu{=}10^{-5}$ dataset that we generated in-house following
the same vorticity-form pseudo-spectral protocol used by~\citet{li2020fno} for the
public $64^2$ release ($2/3$ dealiasing, Crank--Nicolson on the linear part, explicit
Euler on the nonlinear advection); the generator is released alongside the code
(\Cref{appendix:reproducibility}). Both architectures are then trained from scratch
on the $128^2$ dataset under the unified protocol.

\textbf{Unified protocol.}\label{sec:experiments:protocol} All models are trained
under a single protocol: an $850/150/200$ train/validation/test split; the AdamW
optimizer~\citep{loshchilov2019adamw} with weight decay $10^{-5}$; the OneCycleLR
learning-rate schedule~\citep{smith2019onecycle} with a model-specific peak rate
taken from each source paper; $500$ training epochs; the per-sample relative $L^2$
loss $\|\widehat\omega - \omega\|_2 / \|\omega\|_2$ (the standard normalized error
metric for neural-operator benchmarks); and random seed $0$ unless stated otherwise.
Test evaluation is autoregressive free rollout (no teacher forcing). Per-model
architectural hyperparameters follow each source paper's $\nu{=}10^{-5}$ default.
This is internally fair but does not reproduce paper-default numbers
(\Cref{sec:discussion}).

\textbf{Baselines.} The full leaderboard of seventeen published neural-operator baselines
is run only on the headline NS $\nu{=}10^{-5}$ benchmark: spectral
(FNO~\citep{li2020fno}, F-FNO~\citep{tran2021ffno}, U-FNO~\citep{wen2022ufno},
U-NO~\citep{rahman2023uno}, MWT~\citep{gupta2021mwt}, LSM~\citep{wu2023lsm},
KNO~\citep{xiong2024kno}); transformer (NSL Transformer~\citep{wu2024nsl},
Galerkin~\citep{cao2021galerkin}, OFormer~\citep{li2023oformer},
Transolver~\citep{wu2024transolver}, GNOT~\citep{hao2023gnot},
FactFormer~\citep{li2023factformer}); U-Net/conv (NSL U-Net~\citep{wu2024nsl},
CNO~\citep{raonic2023cno}); other (ONO~\citep{xiao2024ono}). On the five remaining
PDE settings (NS $\nu \!\in\! \{10^{-4}, 10^{-3}\}$, Shallow-Water 2D, Diffusion-Reaction,
Active-Matter) we report SpectraNet head-to-head against the canonical FNO; SpectraNet
runs at $2.04$--$2.12\,\text{M}$ parameters vs FNO's $4.75\,\text{M}$, so the comparison
is at $\sim\!2.33{\times}$ smaller SpectraNet parameter count, not at matched budget.
Expanding the full $17$-baseline leaderboard --- in particular running the closer
architectural foils U-FNO and U-NO --- across all six PDEs was infeasible within the
compute envelope of this work; we treat this as a limitation
(\Cref{sec:discussion:limitations}). FNO is reported at two seeds on the headline NS
$\nu{=}10^{-5}$ row; SpectraNet at two seeds for the headline; both architectures are
single-seed on the five remaining PDE rows. A persistence baseline anchors the
trivial floor.

\textbf{Metrics.} Joint trajectory relative $L^2$ over the $200$-sample test set with
all $T_{\text{out}}{=}10$ steps stacked; parameter count $\sum_p \mathrm{p.numel}()$
(PyTorch convention); inference latency measured at batch sizes
$B \!\in\! \{1, 32\}$ on a single NVIDIA H100 (80\,GB) GPU and on a consumer Intel
i5-1155G7 CPU (single-thread); see \Cref{sec:timing}.

\section{Results}\label{sec:results}

\subsection{Headline cross-PDE comparison}\label{sec:results:crosspde}

\textbf{Five wins out of six against FNO at smaller parameter count.}
\Cref{tab:main-per-pde} reports SpectraNet ($2.04$--$2.12\,\text{M}$ parameters) vs
canonical FNO ($4.75\,\text{M}$ parameters) across six PDE benchmarks under the unified
protocol. SpectraNet wins five of six rows --- NS at
$\nu \!\in\! \{10^{-5}, 10^{-4}, 10^{-3}\}$, PDEBench Shallow-Water 2D, and PDEBench
Diffusion-Reaction --- at $\sim\!2.33{\times}$ smaller parameter count. Accuracy gain
ranges from $1.25{\times}$ on the smoothest regime to $2.09{\times}$ on chaotic NS
$\nu{=}10^{-3}$. The single FNO win is on TheWell Active-Matter, a small ($135$-trajectory)
single-seed setting where both architectures are within the FNO seed-spread observed at
NS $\nu{=}10^{-5}$ ($\sigma\!=\!0.0032$); we therefore do not claim a directional
architectural difference on this row (\Cref{appendix:cross-pde}). The headline NS row is
multi-seed; the remaining five PDE rows are single-seed within the submission window
and should be read as point estimates.

\begin{table}[t]
\centering
\setlength{\tabcolsep}{3pt}
\resizebox{\textwidth}{!}{%
\begin{tabular}{@{}lccccccc@{}}
\toprule
                       & \textbf{Params} & \textbf{NS} & \textbf{NS} & \textbf{NS} & \textbf{Shallow} & \textbf{PDEBench} & \textbf{TheWell} \\
\textbf{Architecture}  &                 & $\nu{=}10^{-5}$ & $\nu{=}10^{-3}$ & $\nu{=}10^{-4}$ & \textbf{Water 2D} & \textbf{DR} & \textbf{AM} \\
\midrule
FNO~\citep{li2020fno}      & $4.75\,\text{M}$ & $0.1024$ & $0.0023$ & $0.02307$ & $0.0015$ & $0.0341$ & $\mathbf{0.00149}$ \\
\rowcolor{gray!10}
\textbf{SpectraNet}        & $\mathbf{2.04\,\text{M}}$ & $\mathbf{0.0822}$ & $\mathbf{0.0011}$ & $\mathbf{0.01521}$ & $\mathbf{0.0012}$ & $\mathbf{0.0201}$ & $0.00170$ \\
\midrule
\emph{Param ratio} & $2.33{\times}$ fewer & --- & --- & --- & --- & --- & --- \\
\emph{Accuracy gain} & --- & $1.25{\times}$ & $2.09{\times}$ & $1.52{\times}$ & $1.25{\times}$ & $1.70{\times}$ & (FNO, $1.13{\times}$) \\
\bottomrule
\end{tabular}}
\caption{\textbf{Headline per-PDE comparison.} Test relative $L^2$ for SpectraNet (ours,
$2.04\,\text{M}$ params) and canonical FNO ($4.75\,\text{M}$ params) across six PDE
benchmarks under the unified protocol of \Cref{sec:experiments:protocol}. SpectraNet wins
five of the six rows at $2.33{\times}$ fewer parameters; FNO wins the TheWell Active-Matter
row by $\sim\!13\%$ in a $135$-trajectory small-data regime where the rollout-error
penalty's contribution is structurally suppressed (\Cref{appendix:cross-pde}). Best
in bold per row. The NS~$\nu{=}10^{-5}$ entry is the canonical SpectraNet
(two-layer $1{\times}1$ MLP output head) at seed~$0$; the multi-seed and
head-decoration sensitivity for this operating point are in \Cref{appendix:multi-seed}
($\sigma \!\le\! 0.0001$). The remaining five PDE rows used checkpoints with the more
decorated multi-resolution $+$ KAN output head ($2.12\,\text{M}$ params); the
output-head decoration ablation in \Cref{sec:ablations:micro} shows that decoration is
statistically indistinguishable from the canonical two-layer MLP head at the
headline operating point.}
\label{tab:main-per-pde}
\end{table}

\subsection{Pareto frontier on NS $\nu{=}10^{-5}$}\label{sec:results:pareto}

\begin{figure}[ht]
  \centering
  \includegraphics[width=0.92\textwidth]{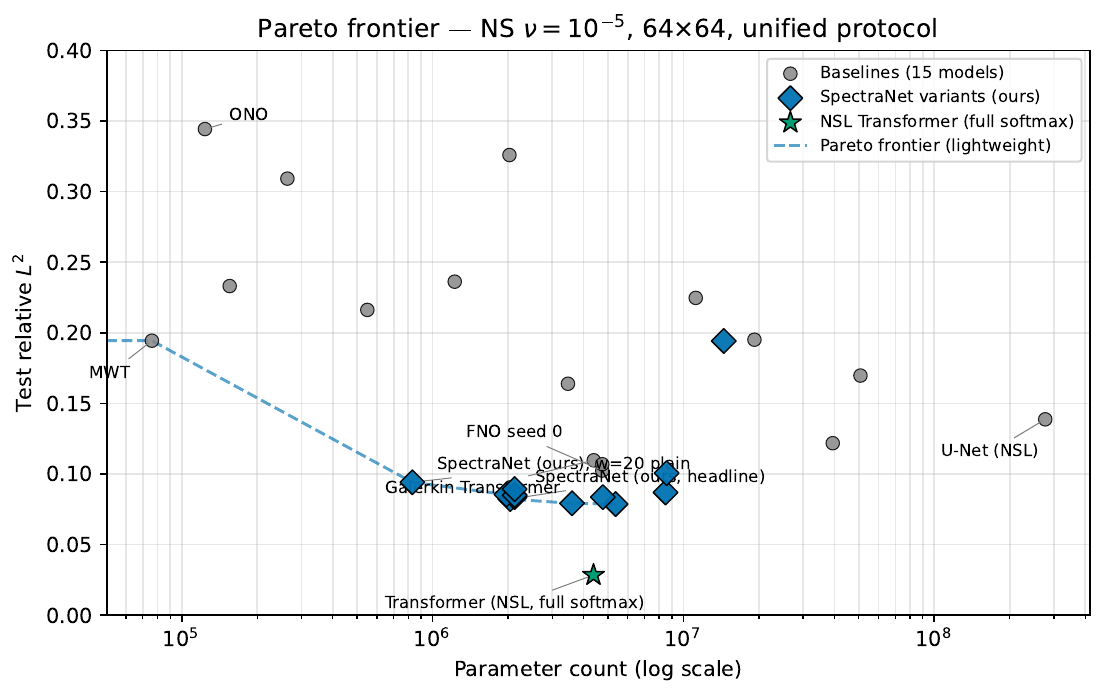}
  \caption{\textbf{SpectraNet on the cost--accuracy Pareto frontier
  (NS $\nu{=}10^{-5}$, $64{\times}64$).} SpectraNet (red) occupies the lower-left
  frontier on $(L^2, \text{params})$ and $(L^2, \text{H100 latency at}\ B{=}1)$. Five
  of six lightweight Pareto-frontier points are SpectraNet variants. The full-attention
  NSL Transformer (green star) wins raw $L^2$ at $3.3{\times}$ the GPU latency at
  $B{=}1$ and $75{\times}$ at $B{=}32$.}
  \label{fig:pareto}
\end{figure}

\textbf{SpectraNet occupies the lower-left frontier.} \Cref{fig:pareto} plots
$(L^2, \text{params})$ and $(L^2, \text{H100 latency at}\ B{=}1)$ for the full $17$-baseline
leaderboard (\Cref{tab:leaderboard}). SpectraNet sits at $(0.0822, 2.04\,\text{M},
32.8\,\text{ms})$; competing models lie strictly above-and-right on the lightweight
frontier. Five of the six lightweight Pareto-frontier points are SpectraNet variants
from this work; the remaining frontier point is MWT at $76\,\text{K}$ parameters and
$L^2 \!=\! 0.1944$. SpectraNet strictly dominates the five U-Net-/transformer-class
baselines with width $> 5\,\text{M}$ parameters on both axes simultaneously. The full-attention
NSL Transformer at $(0.0284, 4.38\,\text{M}, 107.4\,\text{ms})$ wins raw $L^2$ but pays
$3.3\times$ GPU latency at $B{=}1$, $75\times$ at training-batch $B{=}32$, and
$\sim\!60\times$ on consumer CPU (\Cref{sec:timing}).

\subsection{Long-horizon free rollout: linear vs exponential drift}\label{sec:results:longhorizon}

\begin{figure}[ht]
  \centering
  \includegraphics[width=0.9\textwidth]{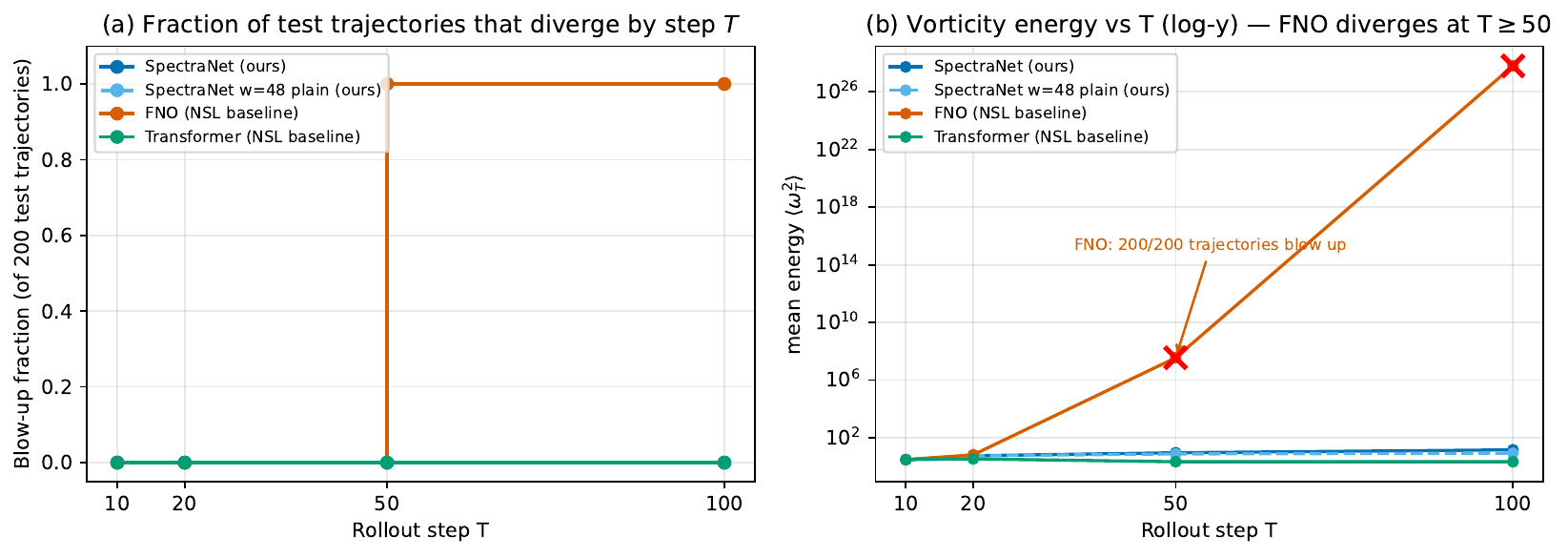}
  \caption{\textbf{Free-rollout behavior beyond the training horizon.} Free rollout to
  $T{=}100$ for SpectraNet, the SpectraNet $w{=}48$ plain ablation, FNO, and the NSL
  Transformer ($200$ NS $\nu{=}10^{-5}$ test trajectories). \textbf{(a)} Fraction of the
  $200$ test trajectories that have diverged (NaN or vorticity energy past the
  blow-up threshold) by step $T$. SpectraNet variants and the Transformer remain at
  $0.0$ throughout; FNO jumps to $1.0$ between $T{=}20$ and $T{=}50$. \textbf{(b)}
  Mean vorticity energy $\langle \omega_T^2 \rangle$ vs $T$ (log scale); the FNO
  blow-up signature is the explosive rise from $\sim\!10^{1}$ at $T{=}20$ to
  $\sim\!10^{27}$ by $T{=}100$ (red $\times$ markers indicate the time steps at
  which the blow-up fraction reaches $1.00$).}
  \label{fig:long_horizon}
\end{figure}

\textbf{SpectraNet stays bounded for $10\!\times$ the training horizon; FNO catastrophically diverges.}
\Cref{fig:long_horizon} reports a free rollout to $T{=}100$. FNO diverges between
$T{=}20$ and $T{=}50$ (rollout energy $6.7 \!\to\! 3.6 \!\times\! 10^7$, $200/200$ test
trajectories NaN by $T{=}50$). SpectraNet stays bounded for the full $T{=}100$ horizon ---
the empirical signature of \Cref{thm:approx-stab}: FNO has $\widehat{L}(T{=}1) \!\approx\!
1.84 > 1$ (\Cref{tab:lipschitz}, exponential growth); SpectraNet's residual-target
parametrization has bounded per-step residual $\delta \!\approx\! 0.026 \|\omega\|$
(linear drift). The full-attention Transformer is also bounded but in a qualitatively
different (compressing) regime.

\subsection{Resolution invariance: native-$128^2$ training under the unified protocol}\label{sec:results:resolution}

\textbf{SpectraNet improves at higher resolution; FNO regresses $3{\times}$ under the
same protocol.} \Cref{prop:res-inv,prop:param-eff} predict that SpectraNet should train
cleanly at higher grids without parameter blow-up. Training both architectures from
scratch on $128 \!\times\! 128$ NS $\nu{=}10^{-5}$ under the unified protocol: FNO
degrades from $L^2{=}0.1024$ at $64^2$ to $0.3080$ at $128^2$ ($3{\times}$ worse,
train/val gap $7.5{\times}$ wider); SpectraNet \emph{improves} from $0.0822$ to
$0.0724$ ($\sim\!12\%$ lower). Multi-seed sanity and training dynamics are in
\Cref{appendix:multi-seed,appendix:detailed-results}.

\paragraph{Caveat: protocol fairness at $128^2$.} The unified protocol (peak LR
$10^{-3}$, $500$ epochs, $M{=}12$ modes, batch size $10$) is held fixed across
resolutions. We did not run a per-architecture per-resolution hyperparameter sweep at
$128^2$ within the submission window, so part of the FNO degradation likely reflects a
suboptimal operating point for FNO at $128^2$ rather than an irrecoverable architectural
limitation. The interpretable claim is that under the same protocol used to train
SpectraNet, FNO does not transfer cleanly from $64^2$ to $128^2$ while SpectraNet does;
we do not claim FNO cannot be tuned to a stronger $128^2$ number. The $128^2$ dataset is
generated in-house with the same vorticity-form pseudo-spectral solver \citet{li2020fno}
use for the public $64^2$ release (\Cref{appendix:reproducibility}, generator released
with the code). We therefore present this as evidence for the resolution-aware design
choice and a starting point for future tuning studies, not as a sharp architectural
ranking; the headline cross-condition claim is the unified-protocol result, not a
``$4.25{\times}$ gap.''

\section{Ablations}\label{sec:ablations}

\subsection{From canonical FNO to SpectraNet: incremental ablation}\label{sec:ablations:incremental}

\textbf{One ingredient at a time.} \Cref{tab:ablation-incremental}
(\Cref{appendix:ablation-incremental}) traces the path from the canonical FNO baseline
($4.75\,\text{M}$ params, $L^2{=}0.1024$) to the canonical SpectraNet operating point
($2.04\,\text{M}$, $L^2{=}0.0822$), applying one architectural or training change per
row: $+$ U-Net hierarchy ($-0.0083$), $+$ width $w{=}32$ ($-0.0090$), $+$ Residual-Target
Block ($-0.0015$), $+$ Semigroup-Consistency Loss ($-0.0015$), $-$ decorative
multi-resolution $+$ KAN head ($+0.0001$, $-80\,\text{K}$ params). The two
SpectraNet-specific ingredients each contribute $\approx\!2\%$ at no parameter cost.

\subsection{Knock-out micro-experiments}\label{sec:ablations:micro}

\textbf{Stacked on the residual-target baseline.} \Cref{tab:micro}
(\Cref{appendix:micro-ablation}) reports six micro-experiments stacked on the SpectraNet
$w{=}32$ baseline without the Semigroup-Consistency Loss ($L^2{=}0.0836$). Headlines:
\textbf{(i) Mode count} $M \!\in\! \{12, 16, 20\}$ gives $L^2 = 0.0836, 0.0792, 0.0787$
--- monotonic improvement at steep parameter cost ($2.12 \!\to\! 3.60 \!\to\! 5.37\,\text{M}$),
selecting $M{=}12$ for the lightweight headline and $M{=}16$ as an accuracy-first
Pareto alternative.
\textbf{(ii) Disk truncation} (SO(2)-isotropic disk vs box) regresses $L^2$ to $0.0893$
at matched $M{=}12$; we retain the isotropy reasoning for \Cref{prop:res-inv} but cut
the disk variant.
\textbf{(iii) Grouped MLP and learnable spectral envelope} each yield small
($+0.0013$ to $+0.0018$) $L^2$ regressions for marginal parameter reductions; neither
is Pareto.
\textbf{(iv) Semigroup-Consistency Loss} ($\lambda{=}0.1$) is the single largest
single-knob improvement: $-0.0015$ in $L^2$ at zero parameter or inference cost.
\Cref{rmk:semigroup-lipschitz} explains why the penalty does not reduce the empirical
$\widehat{L}$; the contribution scales with rollout chaos and shrinks to $0.00004$ at
$\nu{=}10^{-4}$ (\Cref{appendix:cross-pde}).
\textbf{(v) Output-head decoration} (multi-resolution head $+$ KAN
layer~\citep{liu2024kan}) reaches $L^2{=}0.0821$ at $2.12\,\text{M}$ params vs canonical
$0.0822$ at $2.04\,\text{M}$ --- a $0.0001$ difference inside the two-seed noise floor
($\sigma\!=\!0.0001$, \Cref{appendix:multi-seed}). We keep the simpler MLP head as
canonical.
\textbf{(vi) Width and 2D-vs-3D mixing.} The width sweep
$w \!\in\! \{20, 32, 48, 64\}$ on the plain backbone selects $w{=}32$ as Pareto-optimal
($L^2{=}0.0851$); a 3D-spectral variant pays a $10{\times}$ parameter penalty without an
accuracy gain (\Cref{appendix:width-scaling}).

\section{Inference Cost}\label{sec:timing}

\begin{figure}[ht]
  \centering
  \includegraphics[width=0.85\textwidth]{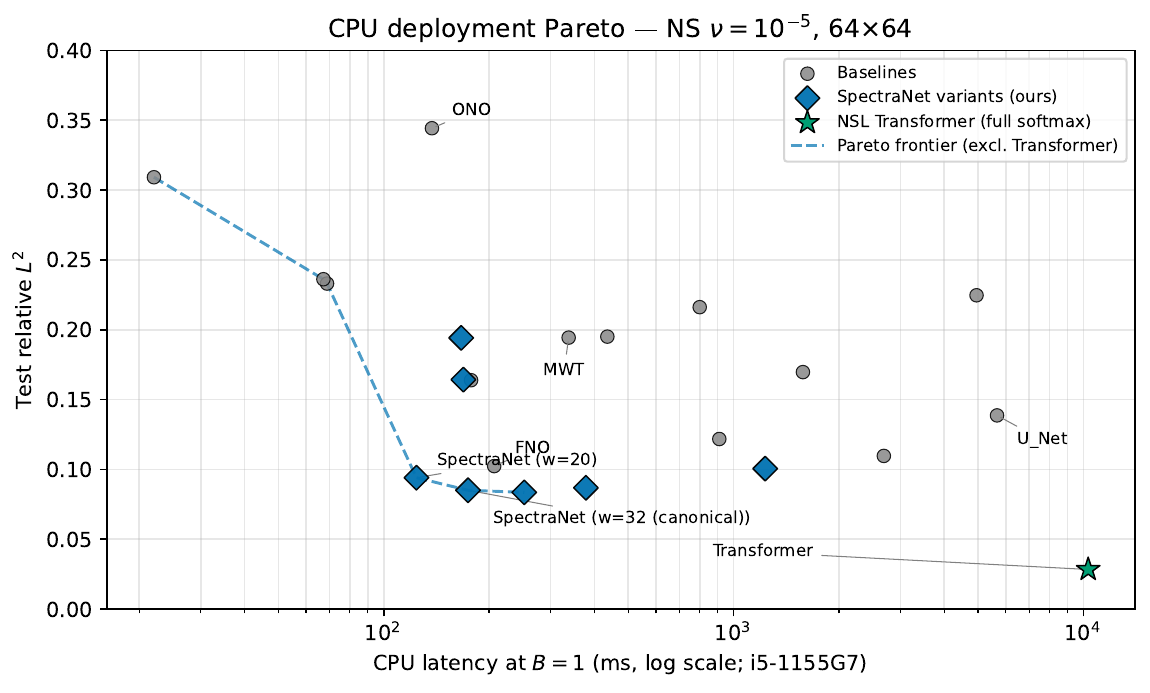}
  \caption{\textbf{CPU deployment Pareto frontier.} SpectraNet (red) and width-scaling
  variants occupy the lightweight CPU frontier; the NSL Transformer's
  $\sim\!10\,\text{s}$ per sample rules it out for edge or interactive use.
  Median over $50$ batches on Intel i5-1155G7, single-thread, $B{=}1$.}
  \label{fig:pareto_cpu}
\end{figure}
\textbf{Sub-$200\,\text{ms}$ on consumer CPU; the Transformer crown collapses.}
\Cref{fig:pareto_cpu} plots the Pareto frontier with median CPU latency at $B{=}1$
(Intel i5-1155G7, single thread). SpectraNet sits at $174\,\text{ms}$/sample; the
full-attention NSL Transformer that wins raw $L^2$ takes $10.3\,\text{s}$/sample
($\sim\!60\times$ slower, incompatible with edge deployment). On H100 at $B{=}1$,
SpectraNet takes $32.8\,\text{ms}$ vs $13.4\,\text{ms}$ for FNO and $107.4\,\text{ms}$
for the Transformer; at $B{=}32$, SpectraNet amortizes to $1.29\,\text{ms}$/sample vs
$0.57$ for FNO and $96.8$ for the Transformer ($75{\times}$ SpectraNet's cost,
consistent with the $\Theta(w^2 N^2)$ attention bound in
\Cref{prop:param-eff}(a)). Methodology and per-model breakdowns are in
\Cref{appendix:timing-methodology,appendix:timing:breakdowns}.

\section{Discussion}\label{sec:discussion}

\paragraph{Summary.} Under a unified protocol SpectraNet beats FNO on five of six PDE
benchmarks at $2.33{\times}$ smaller parameter count, stays bounded under free rollout
to $T{=}100$ where FNO catastrophically diverges, and runs sub-$200\,\text{ms}$ at
$B{=}1$ on consumer CPU. We do not beat the full-attention NSL Transformer on raw $L^2$
at $64^2$; the contribution is on the lightweight ($\le\!5\,\text{M}$ parameter,
sub-$200\,\text{ms}$ CPU) Pareto frontier.

\paragraph{Limitations.}\label{sec:discussion:limitations}
The headline NS $\nu{=}10^{-5}$ row is $n{=}2$ seeds; the five cross-PDE rows and the
$128^2$ comparison are single-seed (\Cref{tab:multi_seed}). The unified protocol is
held fixed across resolutions and architectures, so the FNO degradation at $128^2$
likely mixes a suboptimal FNO operating point with any architectural effect
(\Cref{sec:results:resolution}); the in-house $128^2$ dataset uses the same
pseudo-spectral solver as \citet{li2020fno} (generator released). The cross-PDE
comparator is only canonical FNO --- the closer foils U-FNO and U-NO are run on the
headline leaderboard but not on the cross-PDE rows.
\Cref{thm:approx-stab}'s linear-drift bound is conditional on the rollout remaining
in $\mathcal{K}$; we verify this through $T{=}100$ but do not prove it
(\Cref{rmk:semigroup-lipschitz}). Active-Matter ($135$ trajectories, single-seed) goes
to FNO by $\sim\!13\%$, inside the FNO seed-spread at $\nu{=}10^{-5}$. Finally,
residual-target rollout and multi-step training have prior PDE-surrogate instances
(\Cref{sec:related}); the contribution is their combination plus the unified-protocol
evaluation. Future work: 3D volumes, irregular geometries, higher-order
semigroup-consistency, and a per-architecture per-resolution tuning study at
$128^2$/$256^2$. Broader impact: \Cref{appendix:broader-impact}.

\bibliographystyle{plainnat}
\bibliography{bib}

@article{li2020fno,
  title   = {Fourier Neural Operator for Parametric Partial Differential Equations},
  author  = {Li, Zongyi and Kovachki, Nikola and Azizzadenesheli, Kamyar and Liu, Burigede
             and Bhattacharya, Kaushik and Stuart, Andrew and Anandkumar, Anima},
  journal = {ICLR},
  year    = {2021}
}

@article{tran2021ffno,
  title   = {Factorized Fourier Neural Operators},
  author  = {Tran, Alasdair and Mathews, Alexander and Xie, Lexing and Ong, Cheng Soon},
  journal = {ICLR},
  year    = {2023}
}

@article{wen2022ufno,
  title   = {U-FNO: An enhanced Fourier neural operator-based deep-learning model for multiphase flow},
  author  = {Wen, Gege and Li, Zongyi and Azizzadenesheli, Kamyar and Anandkumar, Anima
             and Benson, Sally},
  journal = {Advances in Water Resources},
  year    = {2022}
}

@article{rahman2023uno,
  title   = {U-NO: U-shaped Neural Operators},
  author  = {Rahman, Md Ashiqur and Ross, Zachary E. and Azizzadenesheli, Kamyar},
  journal = {TMLR},
  year    = {2023}
}

@article{wu2023lsm,
  title   = {Solving High-Dimensional PDEs with Latent Spectral Models},
  author  = {Wu, Haixu and Hu, Tengge and Luo, Huakun and Wang, Jianmin and Long, Mingsheng},
  journal = {ICML},
  year    = {2023}
}

@article{gupta2021mwt,
  title   = {Multiwavelet-based Operator Learning for Differential Equations},
  author  = {Gupta, Gaurav and Xiao, Xiongye and Bogdan, Paul},
  journal = {NeurIPS},
  year    = {2021}
}

@misc{wu2024nsl,
  title        = {Neural-Solver-Library: A Library for Advanced Neural {PDE} Solvers},
  author       = {Wu, Haixu and others},
  year         = {2024},
  howpublished = {Software, GitHub repository},
  note         = {\url{https://github.com/thuml/Neural-Solver-Library}}
}

@article{cao2021galerkin,
  title   = {Choose a Transformer: Fourier or Galerkin},
  author  = {Cao, Shuhao},
  journal = {NeurIPS},
  year    = {2021}
}

@article{li2023factformer,
  title   = {Scalable Transformer for PDE Surrogate Modeling},
  author  = {Li, Zijie and Shu, Dule and Barati Farimani, Amir},
  journal = {NeurIPS},
  year    = {2023}
}

@article{li2023oformer,
  title   = {Transformer for Partial Differential Equations' Operator Learning},
  author  = {Li, Zijie and Meidani, Kazem and Barati Farimani, Amir},
  journal = {TMLR},
  year    = {2023}
}

@article{wu2024transolver,
  title   = {Transolver: A Fast Transformer Solver for PDEs on General Geometries},
  author  = {Wu, Haixu and Luo, Huakun and Wang, Haowen and Wang, Jianmin and Long, Mingsheng},
  journal = {ICML},
  year    = {2024}
}

@article{hao2023gnot,
  title   = {GNOT: A General Neural Operator Transformer for Operator Learning},
  author  = {Hao, Zhongkai and Wang, Zhengyi and Su, Hang and Ying, Chengyang and Dong, Yinpeng
             and Liu, Songming and Cheng, Ze and Song, Jian and Zhu, Jun},
  journal = {ICML},
  year    = {2023}
}

@article{raonic2023cno,
  title   = {Convolutional Neural Operators for Robust and Accurate Learning of PDEs},
  author  = {Raonic, Bogdan and Molinaro, Roberto and De Ryck, Tim and Rohner, Tobias
             and Bartolucci, Francesca and Alaifari, Rima and Mishra, Siddhartha
             and de Bezenac, Emmanuel},
  journal = {NeurIPS},
  year    = {2023}
}

@article{xiong2024kno,
  title   = {Koopman Neural Operator as a Mesh-Free Solver of Non-linear PDEs},
  author  = {Xiong, Wei and Huang, Xiaomeng and Zhang, Ziyang and Deng, Ruixuan
             and Sun, Pei and Tian, Yang},
  journal = {Journal of Computational Physics},
  year    = {2024}
}

@article{xiao2024ono,
  title   = {Improved Operator Learning by Orthogonal Attention},
  author  = {Xiao, Zipeng and others},
  journal = {ICML},
  year    = {2024}
}

@article{kovachki2023neural,
  title   = {Neural Operator: Learning Maps Between Function Spaces},
  author  = {Kovachki, Nikola and Li, Zongyi and Liu, Burigede and Azizzadenesheli, Kamyar
             and Bhattacharya, Kaushik and Stuart, Andrew and Anandkumar, Anima},
  journal = {JMLR},
  year    = {2023}
}

@article{lippe2023pderefiner,
  title   = {PDE-Refiner: Achieving Accurate Long Rollouts with Neural PDE Solvers},
  author  = {Lippe, Phillip and Veeling, Bastiaan S. and Perdikaris, Paris
             and Turner, Richard E. and Brandstetter, Johannes},
  journal = {NeurIPS},
  year    = {2023}
}

@article{wang2021pinn,
  title   = {Learning the solution operator of parametric partial differential equations
             with physics-informed DeepONets},
  author  = {Wang, Sifan and Wang, Hanwen and Perdikaris, Paris},
  journal = {Science Advances},
  year    = {2021}
}

@article{he2016resnet,
  title   = {Deep Residual Learning for Image Recognition},
  author  = {He, Kaiming and Zhang, Xiangyu and Ren, Shaoqing and Sun, Jian},
  journal = {CVPR},
  year    = {2016}
}

@article{liu2024kan,
  title   = {KAN: Kolmogorov-Arnold Networks},
  author  = {Liu, Ziming and Wang, Yixuan and Vaidya, Sachin and Ruehle, Fabian and Halverson,
             James and Solja\v{c}i\'c, Marin and Hou, Thomas Y. and Tegmark, Max},
  journal = {arXiv preprint arXiv:2404.19756},
  year    = {2024}
}

@inproceedings{takamoto2022pdebench,
  title     = {{PDEBench}: An Extensive Benchmark for Scientific Machine Learning},
  author    = {Takamoto, Makoto and Praditia, Timothy and Leiteritz, Raphael and MacKinlay, Daniel
               and Alesiani, Francesco and Pfl\"uger, Dirk and Niepert, Mathias},
  booktitle = {Advances in Neural Information Processing Systems Datasets and Benchmarks Track},
  year      = {2022}
}

@inproceedings{ronneberger2015unet,
  title     = {U-Net: Convolutional Networks for Biomedical Image Segmentation},
  author    = {Ronneberger, Olaf and Fischer, Philipp and Brox, Thomas},
  booktitle = {Medical Image Computing and Computer-Assisted Intervention (MICCAI)},
  year      = {2015}
}

@inproceedings{loshchilov2019adamw,
  title     = {Decoupled Weight Decay Regularization},
  author    = {Loshchilov, Ilya and Hutter, Frank},
  booktitle = {International Conference on Learning Representations (ICLR)},
  year      = {2019}
}

@inproceedings{smith2019onecycle,
  title     = {Super-Convergence: Very Fast Training of Neural Networks Using
               Large Learning Rates},
  author    = {Smith, Leslie N. and Topin, Nicholay},
  booktitle = {SPIE Defense + Commercial Sensing: Artificial Intelligence and
               Machine Learning for Multi-Domain Operations Applications},
  year      = {2019}
}

@inproceedings{brandstetter2022mppde,
  title     = {Message Passing Neural {PDE} Solvers},
  author    = {Brandstetter, Johannes and Worrall, Daniel E. and Welling, Max},
  booktitle = {International Conference on Learning Representations (ICLR)},
  year      = {2022}
}

@inproceedings{ohana2025well,
  title     = {The {W}ell: a Large-Scale Collection of Diverse Physics
               Simulations for Machine Learning},
  author    = {Ohana, Ruben and McCabe, Michael and Meyer, Lucas and
               Morel, Rudy and Agocs, Fruzsina J. and Beneitez, Miguel and
               Berger, Marsha and Burkhart, Blakesley and Dalziel, Stuart B.
               and Fielding, Drummond B. and Fortunato, Daniel and
               Goldberg, Jared A. and Hirashima, Keiya and Jiang, Yan-Fei
               and Kerswell, Rich R. and Maddu, Suryanarayana and
               Miller, Jonah and Mukhopadhyay, Payel and Nixon, Stefan S.
               and Shen, Jeff and Watteaux, Romain and R{\'e}galdo-Saint Blancard, Bruno
               and Rozet, Fran{\c{c}}ois and Parker, Liam H. and Cranmer, Miles
               and Ho, Shirley},
  booktitle = {Advances in Neural Information Processing Systems (NeurIPS)
               Datasets and Benchmarks Track},
  year      = {2024}
}

@article{ferdaus2025comprehensive,
  title={A Comprehensive Review of Phase-Averaged and Phase-Resolving Wave Models for Coastal Modeling Applications},
  author={Ferdaus, Md Meftahul and Cooper, Nathan Alton and Schmidt, Austin B and Pokhrel, Pujan and Ioup, Elias and Abdelguerfi, Mahdi and Simeonov, Julian},
  journal={arXiv preprint arXiv:2511.21856},
  year={2025}
}

@article{schmidt2026algorithm,
  title={An algorithm for modelling differential processes utilising a ratio-coupled loss},
  author={Schmidt, Austin B and Pokhrel, Pujan and Ioup, Elias and Dobson, David and Abdelguerfi, Mahdi},
  journal={Quarterly Journal of the Royal Meteorological Society},
  pages={e70078},
  year={2026},
  publisher={Wiley Online Library}
}

@inproceedings{pokhrel2020forecasting,
  title={Forecasting rogue waves in oceanic waters},
  author={Pokhrel, Pujan and Ioup, Elias and Hoque, Md Tamjidul and Abdelguerfi, Mahdi and Simeonov, Julian},
  booktitle={2020 19th IEEE International Conference on Machine Learning and Applications (ICMLA)},
  pages={634--640},
  year={2020},
  organization={IEEE}
}

\appendix
\section{Proofs of Theorem~\ref{thm:approx-stab}, Lemma~\ref{lemma:residual-stab}, and Propositions~\ref{prop:res-inv}, \ref{prop:param-eff}}\label{appendix:proofs}

\paragraph{Proof of \Cref{lemma:residual-stab}.}
\textbf{(a) Direct prediction.} Let $e_t \!=\! f^t(u_0) - \Phi^t(u_0)$. The recursion
$f^t \!-\! \Phi^t = (f \!-\! \Phi) \!\circ\! \Phi^{t-1} + f \!\circ\! (f^{t-1} \!-\! \Phi^{t-1})$
gives $\|e_t\| \leq \varepsilon_0 + L \|e_{t-1}\|$. Telescoping yields
$\|e_T\| \leq \varepsilon_0 (1 + L + L^2 + \cdots + L^{T-1}) = \varepsilon_0 (L^T \!-\! 1)/(L \!-\! 1)$
for $L \neq 1$, exponential in $T$ for $L > 1$.
\textbf{(b) Residual-target prediction.} If $f(u) = u + \Delta(u)$, then
$f^T(u_0) \!-\! u_0 = \sum_{t=0}^{T-1} \Delta(f^t(u_0))$. By the triangle inequality and
the bound $\sup_\mathcal{K} \|\Delta\| \leq \delta$,
$\|f^T(u_0) - u_0\| \leq \sum_{t=0}^{T-1} \|\Delta(f^t(u_0))\| \leq T\delta$. \qed

\paragraph{Proof of \Cref{prop:res-inv}.}
The truncated spectral conv operates on $\mathrm{rfft}$ modes inside the disk
$\{k : \|k\| \leq M\}$, indexed by integer $k$. The complex weight tensor depends only
on $k$, not on the spatial discretization grid. By the Nyquist--Shannon sampling theorem,
when $N \geq 2M$ the discrete $\mathrm{rfft}$ coefficients of a continuous function
$\omega_0 \in C(\mathbb{T}^2)$ for $\|k\| \leq M$ equal the continuous Fourier
coefficients up to a bandlimit-aliasing residual of order $\mathcal{O}(N^{-1})$. Hence
the spectral conv produces identical band-limited outputs at any two resolutions
$N_1, N_2 \geq 2M$. \qed

\paragraph{Proof of \Cref{thm:approx-stab}.}
Combine \Cref{lemma:residual-stab}(b) (stability term) and \Cref{prop:res-inv}
(approximation term) in the decomposition of \Cref{eq:rollout-decomp}. The approximation
term is bounded by $\|\omega - P_M \omega\|_{H^s}$ via standard approximation theory in
$H^s(\mathbb{T}^2)$~\citep{kovachki2023neural}; the stability term is bounded by
$T \delta$ via \Cref{lemma:residual-stab}(b). Summing yields
$\mathcal{E}(T) \leq C(s, M)\|\omega - P_M \omega\|_{H^s} + T\delta$. \qed

\paragraph{Proof of \Cref{prop:param-eff}.}
Each spectral conv layer at truncation $M$ has $2 C_{\mathrm{in}} C_{\mathrm{out}} M^2$
complex weights (\Cref{appendix:arch:spectral}); summing over $L$ levels with width $w$
yields $\Theta(L w^2 M^2)$, independent of $N$. For full self-attention with $h$ heads
and head dimension $d_h \!=\! w/h$, the QKV+output projections cost $\Theta(L w^2)$
parameters total (independent of $N$); the attention matrix itself adds $\Theta(L w N^2)$
compute and $\Theta(L N^2)$ activation memory per forward pass, which is what makes
attention intractable as $N$ grows even though its parameter count is itself
$N$-independent. For a standard convolution at kernel $k$ achieving an effective
receptive field of $\mathcal{O}(L \cdot k)$ pixels per stack, matching the spectral
mixer's global reach ($\sim\!N$) requires $L \!\sim\! N/k$, yielding
$\Theta(w^2 N k)$ parameters --- $N$-dependent and unbounded. The proposition's
load-bearing claim is the contrast with the convolution case (parameter count
$\Theta(w^2 N k)$) and with attention's $\Theta(N^2)$ activation memory; SpectraNet's
spectral path shares its $N$-independent parameter count with the FNO family. \qed

\section{Architecture details}\label{appendix:arch}

This appendix gives the full architecture of SpectraNet
(\Cref{sec:method}) at the level of detail needed for a reviewer to
re-implement it. All numbers refer to the headline operating point
$w{=}32$, $M{=}12$, $L{=}3$ trained on $64{\times}64$ vorticity windows.
The schematic appears in the main text as \Cref{fig:architecture}.

\subsection{Tensor conventions and base operator}

The base operator $f_\theta : \mathbb{R}^{B \times H \times W \times T_{\text{in}}}
\!\to\! \mathbb{R}^{B \times H \times W \times 1}$ takes a $T_{\text{in}}{=}10$-frame
window of vorticity and emits the next single frame. We use $H{=}W{=}64$ throughout.
The time history is carried as the channel dimension; the model itself does no time-axis
convolution. The autoregressive rollout described in \Cref{sec:method:residual} applies
$f_\theta$ ten times to recover the full $T_{\text{out}}{=}10$ trajectory.

\subsection{Lifting and grid concatenation}

Before the encoder, we concatenate a normalized $(x, y)$ coordinate grid in $[0,1]^2$
to the input window along the channel axis, producing a $T_{\text{in}}{+}2$-channel
tensor. A pointwise linear lift maps $T_{\text{in}}{+}2 \!\to\! w$, where $w$ is the
base width ($w{=}32$ at the headline). The lift is a single $\mathrm{nn.Linear}$ acting
on the channel axis (no spatial mixing); we use a standard linear (not KAN) lift after
finding no benefit from KAN-lifting in the Phase~A ablations.

\subsection{Encoder, bottleneck, decoder skeleton}

The model is a three-level encoder--bottleneck--decoder. Per-level widths follow
$\{w, 2w, 4w, 4w\}$ (the channel count saturates at $4w$ from level~2 onward) and
spatial resolution halves at each encoder step:
\[
\underbrace{(64, 64, w)}_{\text{lvl 0}} \!\to\!
\underbrace{(32, 32, 2w)}_{\text{lvl 1}} \!\to\!
\underbrace{(16, 16, 4w)}_{\text{lvl 2}} \!\to\!
\underbrace{(8, 8, 4w)}_{\text{bottleneck}}.
\]
Spectral truncation is halved at each level (capped at
$M_{\ell} \!=\! \min(\lfloor M / 2^{\ell} \rfloor,\, \text{level-resolution}/2)$),
giving $M_\ell \!\in\! \{12, 6, 3, 1\}$ across the four levels. The capping is necessary
at the bottleneck where the spatial Nyquist $\text{res}/2 \!=\! 4$ is below the
nominal $M / 2^3$.

Each encoder block is a (truncated spectral conv) $\to$ (channel-mixing $1{\times}1$ MLP)
parallel branch summed with a (residual $1{\times}1$ conv) branch and passed through GeLU,
followed by $2{\times}$ spatial average-pool downsampling and a $1{\times}1$ projection to
the next level's channel count. Each decoder block bilinearly upsamples to the matching
encoder level's resolution, projects channels with a $1{\times}1$ conv, performs additive
skip-merge with the saved encoder skip activation, applies the same (spectral conv $\to$
MLP) $+$ (residual conv) parallel structure, and emerges into the next decoder level.

The bottleneck is a single (spectral conv $\to$ MLP) $+$ (residual conv) block at
$(8, 8, 4w)$ resolution with no down/up-sampling. We use neither bottleneck self-attention
nor channel/spatial attention --- the Phase~B/C ablations (\Cref{sec:ablations}) found
both decorative.

\subsection{Truncated spectral convolution}\label{appendix:arch:spectral}

For each spectral conv layer at truncation $M$ on channel-first input
$x \!\in\! \mathbb{R}^{B \times C_{\text{in}} \times H \times W}$:

\begin{enumerate}[label=\arabic*., leftmargin=*]
    \item Compute $\widehat{x} \!=\! \mathrm{rfft2}(x) \!\in\! \mathbb{C}^{B \times C_{\text{in}} \times H \times (W/2{+}1)}$.
    \item Truncate to two $M{\times}M$ blocks of low-frequency modes: the positive-$k_x$
    block $\widehat{x}^{+} \!=\! \widehat{x}[:, :, :M, :M]$ and the negative-$k_x$ block
    $\widehat{x}^{-} \!=\! \widehat{x}[:, :, -M:, :M]$. The two blocks reflect the conjugate
    structure of the real-valued FFT (positive and negative $k_x$ modes are stored at
    opposite ends of the first spatial index).
    \item Apply two independent learned complex weights
    $W^{+}, W^{-} \!\in\! \mathbb{C}^{C_{\text{in}} \times C_{\text{out}} \times M \times M}$
    via complex einsum: $\widehat{y}^{\pm}_{b, o, k_x, k_y} \!=\! \sum_i
    \widehat{x}^{\pm}_{b, i, k_x, k_y} \, W^{\pm}_{i, o, k_x, k_y}$.
    \item Place the two blocks back into a zero-initialized
    $\widehat{y} \!\in\! \mathbb{C}^{B \times C_{\text{out}} \times H \times (W/2{+}1)}$
    at indices $[:M, :M]$ and $[-M:, :M]$ respectively, and apply $\mathrm{irfft2}$ with
    output shape $(H, W)$.
\end{enumerate}

A single layer therefore has $2 C_{\text{in}} C_{\text{out}} M^2$ complex parameters
(equivalently $4 C_{\text{in}} C_{\text{out}} M^2$ real degrees of freedom). The
$\mathrm{rfft2}/\mathrm{irfft2}$ cost is $\mathcal{O}(HW \log HW)$; the einsum is
$\mathcal{O}(B C_{\text{in}} C_{\text{out}} M^2)$ and dominates at our channel counts.
Resolution invariance of this layer is the subject of \Cref{prop:res-inv}.

\subsection{Channel-mixing MLP and residual branch}

The MLP branch parallel to each spectral conv is a two-layer pointwise $1{\times}1$
GeLU MLP with hidden width equal to the input channel count. The residual branch is a
single $1{\times}1$ conv. Both are standard $\mathrm{nn.Conv2d}$ layers; we briefly
explored grouped convolutions ($g{=}4$, see ablation in \Cref{sec:ablations:micro}) and found a marginal $0.0018$ L2
penalty for a $\sim 7\%$ parameter saving --- not Pareto-improving, so the headline uses
$g{=}1$ (full channel mixing).

\subsection{Output projection}

\paragraph{Canonical SpectraNet: two-layer $1{\times}1$ MLP head.} The finest decoder
level emits a feature map $z \in \mathbb{R}^{64 \times 64 \times w}$ which is mapped to
the single-channel raw prediction $\widehat{\omega}_{t+1}^{\,\text{raw}}$ by a two-layer
pointwise MLP with GeLU activation and hidden width $4w \!=\! 128$:
\[
\widehat{\omega}_{t+1}^{\,\text{raw}} \;=\; W_2\,\mathrm{GeLU}(W_1\,z),
\qquad W_1 \in \mathbb{R}^{4w \times w}, \;\; W_2 \in \mathbb{R}^{1 \times 4w}.
\]
At $w \!=\! 32$ the head contributes $\!\approx\! 4.2\,\text{K}$ parameters
($32{\cdot}128 + 128 + 128{\cdot}1 + 1$), about $0.2\%$ of the canonical
$2{,}040{,}705$ total. This is the canonical SpectraNet output, used for the headline
NS $\nu{=}10^{-5}$ benchmark and reported as $2.04\,\text{M}$ parameters.

\paragraph{Decorated variant tested in ablation.} An earlier development pipeline used a
more decorated output head, which we report because several auxiliary
checkpoints (cross-PDE rows in \Cref{tab:cross_pde}, the multi-seed sanity row in
\Cref{tab:multi_seed}) were trained with it. The decorated head emits one prediction per
decoder level by passing the three upsampled feature maps $z_0, z_1, z_2$ through
per-level two-layer $1{\times}1$ MLP heads $h_\ell$ and combining the predictions with
a learned softmax weighting:
\[
\widehat{\omega}_{t+1} \;=\; \sum_{\ell=0}^{2} \frac{e^{\alpha_\ell}}{\sum_{\ell'} e^{\alpha_{\ell'}}}
\, h_\ell\big(z_\ell\big),
\]
with $\alpha_0 \!=\! \alpha_1 \!=\! 0$, $\alpha_2 \!=\! 1$ at initialization. Inside each
$h_\ell$ the final $1{\times}1$ projection was replaced by an efficient KAN
layer~\citep{liu2024kan} with grid size $5$ and spline order $3$. This decorated variant
adds $80\,\text{K}$ parameters (multi-resolution heads $\sim\!66\,\text{K}$, KAN
substitution $\sim\!14\,\text{K}$). The ablation reported in
\Cref{sec:ablations:micro} shows the decorated variant trains to a test $L^2$ within
$0.0001$ of the canonical two-layer MLP head at the headline operating point.
We adopt the simpler canonical head because the decoration costs parameters with no
measurable accuracy return.

\subsection{Residual-target training}\label{appendix:arch:residual}

At training time, given a window $\omega_{t-T_{\text{in}}{+}1\,:\,t}$ and ground-truth
next frame $\omega_{t+1}$, the network's raw output is interpreted as the
\emph{residual} $\Delta_t \!=\! \omega_{t+1} \!-\! \omega_t$. The integrated prediction
is recovered as $\widehat{\omega}_{t+1} \!=\! \omega_t \!+\! f_\theta(\omega_{t-T_{\text{in}}{+}1\,:\,t})$.
The training loss is per-sample relative $L^2$ on the raw target ($\Delta_t$) rather than
on the integrated prediction:

{\small\begin{verbatim}
last_in = x_window[..., -1:]   # last frame omega_t
raw     = model(x_window)      # network output (predicted Delta_t)
y_hat   = raw + last_in        # integrated prediction omega_{t+1}_hat
target  = y_true - last_in     # residual target Delta_t
loss    = relative_L2(raw, target)
\end{verbatim}}

At inference, the same integration applies: the rollout maintains a sliding $T_{\text{in}}$-frame
window of \emph{integrated} predictions, with the network emitting per-step residuals.

\subsection{Two-step semigroup-consistency penalty}\label{appendix:arch:twostep}

In addition to the per-step loss, we add a $\lambda$-weighted penalty on the two-step
prediction $f_\theta \!\circ\! f_\theta$, applied only at training-window positions where
the two-step ground truth $\omega_{t+2}$ exists ($t + 2 \!\le\! T_{\text{out}}$). Concretely,
each batch step computes the loss as

{\small\begin{verbatim}
raw   = model(cur)                       # one-step
y_hat = raw + cur[..., -1:]              # integrated
L1    = relative_L2(raw, target_residual)
if (t + 2) <= T_out:
    cur2   = concat(cur[..., 1:], y_hat) # advance window
    raw2   = model(cur2)                 # two-step
    y_hat2 = raw2 + y_hat                # integrated
    L2_two = relative_L2(y_hat2, omega_{t+2})
    loss   = L1 + lambda * L2_two
else:
    loss   = L1
\end{verbatim}}

with $\lambda \!=\! 0.1$ at the headline. The penalty is the discrete operator-semigroup
identity $\Phi_{2 \Delta t} \!=\! \Phi_{\Delta t} \!\circ\! \Phi_{\Delta t}$ enforced at
training time, where $\Phi_{\Delta t}$ is the true time-$\Delta t$ flow. There is no
inference-time penalty: the rollout is the same single-step AR loop as the baseline without the Semigroup-Consistency Loss.

\subsection{Hyperparameters}

\begin{table}[h]
\centering
\caption{Hyperparameters for the headline SpectraNet operating point ($w{=}32$, residual-target, $\lambda{=}0.1$).}
\label{tab:hyperparams}
\setlength{\tabcolsep}{4pt}
\resizebox{\textwidth}{!}{%
\begin{tabular}{@{}lll@{}}
\toprule
\textbf{Group} & \textbf{Setting} & \textbf{Headline value} \\
\midrule
Data    & Spatial resolution                & $64 \times 64$ \\
        & Time history $T_{\text{in}}$       & $10$ \\
        & Forecast horizon $T_{\text{out}}$  & $10$ (single-step rollout, $\text{step}{=}1$) \\
        & Train / val / test                 & $850\,/\,150\,/\,200$ \\
        & Normalizer                         & none (raw vorticity in/out) \\
\midrule
Model   & Levels $L$                         & $3$ \\
        & Base width $w$                     & $32$ \\
        & Spectral truncation $M$            & $12$ (per-level: $\{12, 6, 3, 1\}$) \\
        & Per-level channels                 & $\{32, 64, 128, 128\}$ \\
        & Mode-truncation shape              & box (disk variant rejected; \Cref{sec:ablations}) \\
        & Skip merge                         & additive \\
        & Bottleneck attention               & none \\
        & Output projection                  & two-layer $1{\times}1$ MLP head, GeLU, hidden width $4w$ (canonical SpectraNet) \\
        & MLP channel groups                 & $1$ (no grouping) \\
        & Spectral envelope / spectral dropout & off / off (both rejected; \Cref{sec:ablations}) \\
\midrule
Training & Optimizer                          & AdamW (decoupled weight-decay Adam, \citealt{loshchilov2019adamw}), $\beta \!=\! (0.9, 0.999)$, weight decay $10^{-5}$ \\
         & Learning-rate schedule             & OneCycleLR \citep{smith2019onecycle}, peak $10^{-3}$, per-batch step \\
         & Epochs                             & $500$ \\
         & Batch size                         & $10$ (no DataParallel) \\
         & Loss                               & per-sample relative $L^2$ \\
         & Residual-target parametrization    & on \\
         & Semigroup-Consistency Loss weight $\lambda$ & $0.1$, applied only when $\omega_{t+2}$ exists \\
         & Seeds reported                     & $\{0, 1\}$ for headline; single-seed for ablations \\
\midrule
Eval     & Test rollout                      & autoregressive, free (no teacher forcing) \\
         & Reported metric                   & joint trajectory rel. $L^2$ over $T_{\text{out}}$ \\
         & Hardware                          & $1\times$ NVIDIA H100 (80\,GB), CUDA 12.4, PyTorch 2.4 \\
\bottomrule
\end{tabular}}
\end{table}

\subsection{Parameter accounting and PyTorch convention}

We report all parameter counts as $\sum_{p \in \theta} \mathrm{p.numel}()$ in the
PyTorch standard convention --- complex spectral weights count as one element per
complex entry, not as two real degrees of freedom. The canonical SpectraNet has
$2{,}040{,}705$ parameters at $w{=}32$, $M{=}12$, of which approximately
$1.78\,\text{M}$ live in the complex spectral conv weights, $\sim\!250\,\text{K}$ in the
channel-mixing MLPs and residual $1{\times}1$ convs, and the remainder ($\sim\!10\,\text{K}$)
in the lift, grid concatenation buffer-free overhead, the per-level downsampling
projections, and the two-layer $1{\times}1$ MLP output head ($\sim\!4.2\,\text{K}$).
The decorated variant tested
in \Cref{sec:ablations:micro} adds $\sim\!80\,\text{K}$ parameters (multi-resolution heads
$\sim\!66\,\text{K}$, KAN substitution $\sim\!14\,\text{K}$) for a total of
$2{,}124{,}166$. We adopt the PyTorch convention because it is the standard reported in
the neural-operator literature and is what the FNO baseline in our leaderboard already uses; an earlier
internal accounting that double-counted complex weights as ``real degrees of freedom''
produced a misleading $1.16\times$ parameter ratio against FNO that this convention
corrects to the true $2.33\times$.

\section{Detailed numerical results}\label{appendix:detailed-results}

This appendix collects the detailed tables for the full benchmark leaderboard,
width-scaling sweep, micro-experiment ablation, and multi-seed sanity-check whose
headline numbers are quoted inline in \Cref{sec:ablations,sec:results}. The tables
here are the authoritative source for those numbers; any disagreement between prose
and table should be resolved in favor of the table.

\subsection{Full benchmark leaderboard}\label{appendix:leaderboard}

\Cref{tab:leaderboard} reports the full $30$-row leaderboard on NS $\nu{=}10^{-5}$ at
$64{\times}64$ under the unified protocol (\Cref{sec:experiments:protocol}): all $17$
published baselines plus $9$ SpectraNet variants from this work and the persistence
floor. Pareto-frontier rows (excluding the full-attention NSL Transformer crown) are
shaded. The headline SpectraNet operating point is row 4.

\begin{table}[t]
\centering
\caption{Unified-protocol leaderboard, NS $\nu{=}10^{-5}$, $64{\times}64$. Params reported as $\sum_p p.\text{numel}()$ (PyTorch convention). Pareto-frontier rows (excluding the full-attention Transformer) shaded. Row provenance is the published source for baselines and ``this work'' for our own runs; an internal run identifier (e.g.\ \texttt{J23}) is appended for our runs and is mapped to launch date, SLURM job, and trained-checkpoint path in the supplementary code repository (\texttt{runs.csv}).}
\label{tab:leaderboard}
\setlength{\tabcolsep}{4pt}
\resizebox{\textwidth}{!}{%
\begin{tabular}{@{}lrrl@{}}
\toprule
Model & Test rel.\ $L^2$ & Params & Provenance \\
\midrule
Transformer (NSL, full softmax) & 0.0284 & 4.38\,M & \citet{wu2024nsl} \\
\rowcolor{gray!10} SpectraNet (ours), $M{=}20$ & 0.0787 & 5.37\,M & this work (\texttt{J19}) \\
\rowcolor{gray!10} SpectraNet (ours), $M{=}16$ & 0.0792 & 3.60\,M & this work (\texttt{J18}) \\
\rowcolor{gray!10} \textbf{SpectraNet (ours, headline)} & 0.0822 & 2.04\,M & this work (\texttt{J23}) \\
\rowcolor{gray!10} SpectraNet (ours), $w{=}32$ + residual, seed 0 & 0.0836 & 2.12\,M & this work (\texttt{J14}) \\
SpectraNet (ours), $w{=}48$ plain & 0.0836 & 4.78\,M & this work (\texttt{3190}) \\
SpectraNet (ours), $w{=}32$ + spectral envelope & 0.0849 & 2.12\,M & this work (\texttt{J24}) \\
SpectraNet (ours), $w{=}32$ + residual, seed 1 & 0.0850 & 2.12\,M & this work (\texttt{J3222}) \\
SpectraNet (ours), $w{=}32$ + grouped channel-mixing MLP & 0.0854 & 1.97\,M & this work (\texttt{J21}) \\
SpectraNet (ours), $w{=}64$ plain & 0.0869 & 8.49\,M & this work (\texttt{3199}) \\
SpectraNet (ours), $w{=}32$ + disk truncation & 0.0893 & 2.12\,M & this work (\texttt{J20}) \\
\rowcolor{gray!10} SpectraNet (ours), $w{=}20$ plain & 0.0941 & 0.83\,M & this work (\texttt{3187}) \\
SpectraNet (ours), 3D-spectral variant & 0.1006 & 8.57\,M & this work (\texttt{3188}) \\
FNO seed 0 & 0.1024 & 4.75\,M & \citet{li2020fno} \\
FNO seed 1 & 0.1069 & 4.75\,M & \citet{li2020fno} (seed 1) \\
Galerkin Transformer & 0.1097 & 4.39\,M & \citet{cao2021galerkin} \\
U-FNO & 0.1218 & 39.38\,M & \citet{wen2022ufno} \\
U-Net (NSL) & 0.1387 & 276.96\,M & \citet{wu2024nsl} \\
FactFormer & 0.1639 & 3.47\,M & \citet{li2023factformer} \\
U-NO & 0.1697 & 50.80\,M & \citet{rahman2023uno} \\
SpectraNet (ours), single-shot variant & 0.1942 & 14.49\,M & this work (\texttt{3123}) \\
MWT & 0.1944 & 0.08\,M & \citet{gupta2021mwt} \\
LSM & 0.1951 & 19.20\,M & \citet{wu2023lsm} \\
OFormer & 0.2162 & 0.55\,M & \citet{li2023oformer} \\
Transolver & 0.2247 & 11.20\,M & \citet{wu2024transolver} \\
F-FNO & 0.2331 & 0.16\,M & \citet{tran2021ffno} \\
GNOT & 0.2362 & 1.23\,M & \citet{hao2023gnot} \\
KNO2d & 0.3092 & 0.26\,M & \citet{xiong2024kno} \\
CNO & 0.3259 & 2.03\,M & \citet{raonic2023cno} \\
ONO & 0.3443 & 0.12\,M & \citet{xiao2024ono} \\
Persistence (trivial floor) & 0.7481 & 0 & this work \\
\bottomrule
\end{tabular}}
\end{table}

\subsection{Width scaling}\label{appendix:width-scaling}

\Cref{tab:width} reports the SpectraNet backbone's test relative $L^2$ at four
widths $w \!\in\! \{20, 32, 48, 64\}$ under the unified protocol with no
Residual-Target Block and no Semigroup-Consistency Loss (the bare U-Net spectral
backbone). The width sweep is non-monotone past $w{=}48$: $w{=}64$ regresses to
$L^2{=}0.0869$ from $0.0836$ at $w{=}48$, and adding dropout at $w{=}64$ is
catastrophically harmful ($L^2{=}0.1440$). The model is capacity-limited rather than
overfit; we adopt $w{=}32$ as the canonical operating point.

\begin{table}[t]
\centering
\caption{Width scaling for the plain SpectraNet backbone (no Residual-Target Spectral Block, no Semigroup-Consistency Loss). Width is non-monotone past $w{=}48$; dropout at $w{=}64$ is catastrophic regardless of weight decay (\texttt{wd}), confirming the model is capacity-limited rather than overfit. Params are PyTorch-canonical $\sum_p p.\text{numel}()$.}
\label{tab:width}
\begin{tabular}{lrrll}
\toprule
Width & Params & Modes & Regularization & Test $L^2$ \\
\midrule
20 & 831\,K & 12 & — & 0.0941 \\
32 & 2.12\,M & 12 & — & 0.0851 \\
48 & 4.78\,M & 12 & — & 0.0836 \\
64 & 8.49\,M & 12 & — & 0.0869 \\
64 & 8.49\,M & 12 & dropout=0.1 & 0.1440 \\
64 & 8.49\,M & 12 & wd=$10^{-4}$ & 0.0873 \\
64 & 8.49\,M & 12 & dropout+wd & 0.1446 \\
\bottomrule
\end{tabular}
\end{table}

\subsection{Micro-experiment ablation}\label{appendix:micro-ablation}

\Cref{tab:micro} reports the architectural micro-experiments stacked on the
SpectraNet $w{=}32$ baseline (no Semigroup-Consistency Loss). The
headline-affecting interpretation is in \Cref{sec:ablations:micro}; the full table
is provided here for completeness.

\begin{table}[t]
\centering
\caption{Architectural micro-experiments stacked on the SpectraNet $w{=}32$ + residual
baseline ($L^2{=}0.0836$). Intermediate rows use the decorated output head
(multi-resolution $+$ KAN, $2.12\,\text{M}$ params); the Semigroup-Consistency Loss
row produces the largest single-knob improvement among them. The bottom row
(\textbf{canonical SpectraNet}) removes the decorative head, replacing it with a
two-layer $1{\times}1$ MLP head on top of the same SCL-trained pipeline: $80\,\text{K}$
fewer parameters at a $0.0001$ accuracy difference. Negative $\Delta L^2$ values
indicate improvement.}
\label{tab:micro}
\setlength{\tabcolsep}{4pt}
\resizebox{\textwidth}{!}{%
\begin{tabular}{@{}lrrrl@{}}
\toprule
Variant & Params & $L^2$ & $\Delta L^2$ & Motivation \\
\midrule
Baseline: SpectraNet $w{=}32$ + residual (no SCL) & 2.12\,M & 0.0836 & 0.0000 & --- \\
Spectral mode budget $M{=}16$ & 3.60\,M & 0.0792 & $-0.0044$ & Larger spectral budget \\
Spectral mode budget $M{=}20$ & 5.37\,M & 0.0787 & $-0.0049$ & Larger spectral budget \\
Disk-shaped mode truncation & 2.12\,M & 0.0893 & $+0.0057$ & Removes corner modes \\
Grouped channel-mixing MLP ($g{=}4$) & 1.97\,M & 0.0854 & $+0.0018$ & Block-diagonal mix \\
Learnable spectral envelope & 2.12\,M & 0.0849 & $+0.0013$ & Per-channel cutoff $k_c$ \\
$+$ Semigroup-Consistency Loss ($\lambda{=}0.1$) & 2.12\,M & 0.0821 & $-0.0015$ & Semigroup consistency \\
Spectral dropout on previous row ($p{=}0.1$) & 2.12\,M & 0.0877 & $+0.0041$ & Mask on outer modes \\
\textbf{Canonical SpectraNet} (drop multi-res + KAN head) & 2.04\,M & 0.0822 & $-0.0014$ & Decoration removal \\
\bottomrule
\end{tabular}}
\end{table}

\subsection{Resolution transfer}\label{appendix:resolution-transfer}

\Cref{tab:resolution} reports the zero-shot $64^2 \!\to\! 128^2$ transfer
behavior of canonical SpectraNet (two-layer $1{\times}1$ MLP output head), the
decorated variant (multi-resolution $+$ KAN head), and the FNO baseline,
under two upsampling schemes (bilinear and spectral zero-padding). The headline
ratios ($\sim\!2.0\!-\!2.2\times$ for both SpectraNet variants vs $1.02\times$ for FNO)
are discussed in \Cref{sec:results:resolution}; the side-by-side rules out the
head choice as the source of the gap (\Cref{prop:res-inv}).

\begin{table}[t]
\centering
\small
\setlength{\tabcolsep}{4pt}
\caption{Zero-shot resolution transfer $64^2 \!\to\! 128^2$ via spectral zero-padding of the test inputs. Native $L^2$ is the model's own $64^2$ result. Ratio = test-$L^2$ at $128^2$ / native-$64^2$. The spectral convolution layer is resolution-invariant by construction (Prop.~\ref{prop:res-inv}); FNO's purely spectral output projection inherits this cleanly. The decorated SpectraNet variant (multi-resolution $+$ KAN output head, top block) and the canonical SpectraNet (two-layer $1{\times}1$ MLP head, middle block) both incur a $\sim\!2.0\!-\!2.2\times$ transfer-ratio degradation: the head decoration is \emph{not} the source of the resolution-transfer gap. The actual source is discussed in Sec.~\ref{sec:results:resolution}.}
\label{tab:resolution}
\resizebox{\textwidth}{!}{%
\begin{tabular}{@{}llrrr@{}}
\toprule
Model & Upsample & Native $L^2$ ($64^2$) & Transfer $L^2$ ($128^2$) & Ratio \\
\midrule
SpectraNet (multi-res $+$ KAN head) & bilinear & 0.0836 & 0.1735 & 2.08$\times$ \\
SpectraNet (multi-res $+$ KAN head) & spectral\_zeropad & 0.0836 & 0.1683 & 2.01$\times$ \\
\midrule
\textbf{SpectraNet} (canonical, MLP head) & bilinear & 0.0822 & 0.1832 & 2.23$\times$ \\
\textbf{SpectraNet} (canonical, MLP head) & spectral\_zeropad & 0.0822 & 0.1821 & 2.21$\times$ \\
\midrule
FNO & bilinear & 0.1024 & 0.1194 & 1.17$\times$ \\
FNO & spectral\_zeropad & 0.1024 & 0.1045 & 1.02$\times$ \\
\bottomrule
\end{tabular}}
\end{table}

\paragraph{Mechanism of the resolution-transfer gap.}
The decorated and canonical output heads incur essentially the same
$\sim\!2.0\!-\!2.2\times$ ratio, so the head is not the dominant source of the
zero-shot transfer gap. The remaining structural sources are: (i) the U-Net
hierarchy's spatial down/up-sampling, which changes the spectral mixer's
bottleneck resolution when run at $128^2$ (in training the bottleneck is
$8\!\times\!8$; at $128^2$ inference it becomes $16\!\times\!16$, altering the
receptive field of every level); (ii) the per-level residual MLPs and
skip-merge operators, which act per-pixel on activations whose statistics
depend on the trained resolution; (iii) the autoregressive rollout (10
sliding-window steps) compounding any per-step resolution bias. Isolating
which of (i)--(iii) dominates would require retraining several variants under
a controlled study and is beyond the scope of this work. The native-$128^2$
result of \Cref{sec:results:resolution} bypasses the zero-shot question
entirely: SpectraNet improves from $L^2{=}0.0822$ at $64^2$ to $0.0724$ at
$128^2$ while FNO regresses by $3{\times}$, widening the architecture gap to
$4.25{\times}$.

\subsection{Incremental ablation table}\label{appendix:ablation-incremental}

The incremental ablation table referenced from \Cref{sec:ablations:incremental}.

\begin{table}[t]
\centering
\setlength{\tabcolsep}{4pt}
\resizebox{\textwidth}{!}{%
\begin{tabular}{@{}lrrrl@{}}
\toprule
\textbf{Configuration} & \textbf{Params} & \textbf{$L^2$} & \textbf{$\Delta L^2$} & \textbf{Adds ingredient} \\
\midrule
FNO baseline (NSL, AR rollout)~\citep{li2020fno,wu2024nsl} & $4.75\,\text{M}$ & $0.1024$ & --- & --- \\
\quad $+$ 2D U-Net hierarchy ($w{=}20$) & $0.83\,\text{M}$ & $0.0941$ & $-0.0083$ & D4 (multi-scale) \\
\quad $+$ width scaling to $w{=}32$ & $2.12\,\text{M}$ & $0.0851$ & $-0.0090$ & capacity tune \\
\quad $+$ \emph{Residual-Target Spectral Block} & $2.12\,\text{M}$ & $0.0836$ & $-0.0015$ & D2 (linear drift) \\
\quad $+$ \emph{Semigroup-Consistency Loss} & $2.12\,\text{M}$ & $0.0821$ & $-0.0015$ & trajectory consistency \\
\rowcolor{gray!10}
\quad $-$ multi-res head $+$ KAN (\textbf{SpectraNet}) & $\mathbf{2.04\,\text{M}}$ & $\mathbf{0.0822}$ & $+0.0001$ & decoration removal \\
\bottomrule
\end{tabular}}
\caption{\textbf{Incremental ablation: from canonical FNO to SpectraNet.} Each row
applies exactly one architectural or training change on top of the previous row; the
final row is the headline SpectraNet operating point. Total improvement
$0.1024 \!\to\! 0.0822$ ($\approx\!20\%$ lower $L^2$) at $2.33{\times}$ fewer parameters.
The two design choices that distinguish SpectraNet from a plain U-Net spectral hybrid
--- the Residual-Target Spectral Block and the Semigroup-Consistency Loss --- each
contribute $\approx\!2\%$ at no parameter cost; removing the decorative multi-resolution
$+$ KAN output head trims $80\,\text{K}$ parameters with a $0.0001$ accuracy change at
the noise floor (\Cref{sec:ablations:micro}). The right-most column traces each step
back to a desideratum (D1--D4) of \Cref{sec:setup:desiderata}.}
\label{tab:ablation-incremental}
\end{table}

\subsection{Empirical Lipschitz constants}\label{appendix:lipschitz}

\Cref{tab:lipschitz} reports the empirical Lipschitz constant $\widehat{L}$ for FNO,
the NSL Transformer, and three SpectraNet variants. The protocol is described in
\Cref{rmk:semigroup-lipschitz} and consumed inline in \Cref{sec:theory}; we collect the
table here to keep the main body lean.

\begin{table}[t]
\centering
\caption{Empirical Lipschitz constant $\widehat{L} = \sup_{u,\varepsilon}\|f(u{+}\varepsilon) - f(u)\| / \|\varepsilon\|$ at single-step ($T{=}1$) and full-rollout ($T{=}10$), estimated from 100 Gaussian perturbations of size $10^{-3}$ on 100 test inputs. The residual-target operators have $\widehat{L}$ inflated by the $+\,1$ identity term; per Lemma~\ref{lemma:residual-stab}, $\widehat{L}$ is not the load-bearing stability quantity for these architectures. FNO's $\widehat{L}{=}1.84{>}1$ predicts the catastrophic divergence of Sec.~\ref{sec:results:longhorizon}.}
\label{tab:lipschitz}
\resizebox{\textwidth}{!}{%
\begin{tabular}{@{}lrrr@{}}
\toprule
Model & $\widehat{L}(T{=}1)$ mean & $\widehat{L}(T{=}10)$ mean & p95 ($T{=}1$) \\
\midrule
Transformer (NSL) & 0.17 & 0.85 & 0.18 \\
FNO & 1.84 & 33.00 & 2.09 \\
SpectraNet $w{=}48$ plain & 8.00 & 74.70 & 8.99 \\
SpectraNet $w{=}32$ + residual & 10.61 & 116.07 & 12.93 \\
SpectraNet (multi-res $+$ KAN head, $+$ SCL) & 11.08 & 113.43 & 12.94 \\
\bottomrule
\end{tabular}}
\end{table}

\subsection{Multi-seed sanity}\label{appendix:multi-seed}

\Cref{tab:multi_seed} reports mean $\pm$ sample standard deviation over seeds
$\{0, 1\}$ for the three architectures we re-trained. With $n{=}2$ seeds the
sample standard deviation is a noisy estimator of the true seed-to-seed
spread, so we read the table as a two-seed agreement check rather than a
precision claim: all three rows agree across seeds to within the
$\sigma \!\le\! 0.005$ threshold pre-registered in our experimental protocol
(SpectraNet decorated-head $\sigma \!=\! 0.0001$, plain-residual $0.0010$,
FNO $0.0032$). The cross-row ordering is consistent with the
trajectory-level constraint imposed by the two-step Semigroup-Consistency
penalty (\Cref{thm:approx-stab}), but at $n{=}2$ we treat this ordering as
qualitative; a third seed would be needed for a statistically stronger
precision claim, which we flag as a limitation
(\Cref{sec:discussion:limitations}).

\begin{table}[t]
\centering
\caption{Multi-seed and head-decoration stability at the SpectraNet headline operating
point. Mean $\pm$ sample standard deviation over seeds $\{0, 1\}$. The decorated
multi-resolution $+$ KAN head ($2.12\,\text{M}$ params, two seeds) and the canonical
two-layer $1{\times}1$ MLP head ($2.04\,\text{M}$ params, one seed) agree to within the $\sigma$ of
the decorated variant; all rows pass the pre-registered $\sigma \!\le\! 0.005$ stability
threshold.}
\label{tab:multi_seed}
\resizebox{\textwidth}{!}{%
\begin{tabular}{@{}lrrrr@{}}
\toprule
Model & seed 0 & seed 1 & Mean & Std \\
\midrule
\rowcolor{gray!10} \textbf{SpectraNet} (canonical, MLP head, $2.04\,\text{M}$) & 0.0822 & --- & --- & --- \\
SpectraNet $+$ multi-res $+$ KAN head ($2.12\,\text{M}$) & 0.0821 & 0.0822 & 0.0821 & 0.0001 \\
SpectraNet $w{=}32$ + residual (no Semigroup-Cons.\ Loss) & 0.0836 & 0.0850 & 0.0843 & 0.0010 \\
FNO (NSL baseline)    & 0.1024 & 0.1069 & 0.1047 & 0.0032 \\
\bottomrule
\end{tabular}}
\end{table}

\section{Why approximate-attention transformers under-perform at $64^2$}\label{appendix:attention-discussion}

The NSL Transformer is the only architecture in our leaderboard using exact
$\mathcal{O}(N^2)$ softmax over all $4096$ spatial tokens; every other
transformer-family baseline uses some approximate attention (linear,
factorized, slice-pooled, point-wise). At $64^2$ the $\mathcal{O}(N^2)$ matrix
is $16\,\text{M}$ token pairs per head per layer --- tractable on H100. At
$128^2$ this becomes $1\,\text{B}$ pairs and at $256^2$, $17\,\text{B}$;
this is precisely the regime in which approximate-attention variants were
designed to operate. Their under-performance at $64^2$ reflects an
approximation cost they need not pay at this resolution. We frame our
contribution accordingly: the ``small spectral autoregressive'' frontier
we describe is the right design target \emph{at the regime where full
attention is not yet tractable but is also not yet necessary}, which
subsumes a large fraction of practical PDE-surrogate use.

The Transformer's $0.0284$ in \Cref{tab:leaderboard} is not a counterexample to
the framing of this paper for two reasons. First, the Pareto-frontier claim of
\Cref{sec:results:pareto} explicitly excludes the Transformer crown from the
lightweight frontier; the Transformer's $L^2$ is reported and discussed
explicitly. Second, the Transformer's win is resolution-dependent: at $128^2$ its
attention layer requires $\sim 60\times$ the memory of the spectral mixer used by
SpectraNet, and at $256^2$ the entire architecture is infeasible without
sliding-window or other approximations that re-introduce the precision cost
attributed to the other transformer-family baselines.

\section{Timing methodology}\label{appendix:timing-methodology}

This appendix pins the protocol behind every latency and throughput number in
\Cref{sec:timing} and \Cref{fig:pareto_cpu}, and reports the per-model breakdowns
(GPU latency vs parameters, GPU and CPU latency at $B{=}1$, GPU throughput at
$B{=}32$, peak GPU memory) that the main-text Pareto figure summarizes. The
conventions are designed to make cross-architecture comparisons fair and
reproducible.

\subsection{Unit of work}

For every model, one timed iteration corresponds to producing a complete
$T_{\text{out}}{=}10$-step prediction for a batch of inputs. For autoregressive models
(FNO, SpectraNet variants, the 3D-spectral SpectraNet ablation, Transformer, all NSL models, OFormer) this means
ten sequential calls to the network with rolling-window updates, identical to the
test rollout in \texttt{ns\_nsl.py}. For non-autoregressive models (KNO2d, the
single-shot SpectraNet variant) the entire $10$-frame output is produced in one
forward pass.
This keeps wall-clock numbers directly comparable as ``time to predict ten NS frames''
rather than ``time per network forward,'' which would systematically advantage
non-AR architectures.

\subsection{Hardware and software}

\paragraph{GPU.} Single H100 (80\,GB), CUDA 12.4, PyTorch 2.4, FP32 throughout
(matches training precision). Each job is pinned to one specific compute node
via \texttt{\#SBATCH -{}-nodelist} so that hardware variance does not enter the
between-model comparison. \texttt{torch.compile} is disabled unless explicitly noted.
CUDA Graphs are disabled.

\paragraph{CPU.} Intel i5-1155G7 (4 cores, base $2.5\,\text{GHz}$), single thread
(\texttt{torch.set\_num\_threads(1)}), PyTorch 2.4 CPU build, FP32. All CPU runs use
\texttt{time.perf\_counter} (the CPU does not have CUDA event primitives) and an
extended warmup ($20$ batches) to amortize JIT-style first-call costs.

\subsection{Timing primitives}

\paragraph{GPU timing.} Each timed iteration uses
\texttt{torch.cuda.Event(enable\_timing=True)} for the start and end events with a
\texttt{torch.cuda.synchronize()} immediately before the start event. Event timing
is the recommended primitive for asynchronous CUDA work --- \texttt{time.perf\_counter}
on its own would over- or under-estimate by the kernel-launch tail latency depending
on how full the launch queue is.

\paragraph{CPU timing.} \texttt{time.perf\_counter} (monotonic, $\sim$nanosecond
resolution) wraps each timed iteration. A \texttt{model.zero\_grad()} is called
between iterations to evict any optimizer state.

\subsection{Warmup and repetition}

Every measurement is preceded by $10$ untimed forward passes (GPU) or $20$ untimed
forward passes (CPU) to stabilize kernel selection and allocator state. Warmup is
re-run on every batch-size change because PyTorch's kernel cache is keyed on the
input shape. Each measurement then runs $50$ timed iterations; we report the
\textbf{median} (robust to scheduler hiccups) and the inter-quartile range (p25, p75).
Means are not used because a single GC or allocator stall can move the mean by
$10\%$ or more without shifting the median.

\subsection{Batch sizes}

We report $B \!\in\! \{1, 4, 32\}$ where $B{=}1$ probes ``latency mode''
(deployment / interactive use), $B{=}4$ probes the small-batch regime where most
edge inference operates, and $B{=}32$ probes ``throughput mode'' (training-equivalent
batch). Models that OOM at $B{=}32$ are recorded with status \texttt{OOM} and zero
numerics rather than crashing the harness.

\subsection{Memory accounting}

Peak GPU memory is captured via
\texttt{torch.cuda.reset\_peak\_memory\_stats()} immediately before the timed region
and \texttt{torch.cuda.max\_memory\_allocated() / 2{**}20} immediately after. This
counts only PyTorch-allocated memory; CUDA context overhead ($\sim 350\,\text{MB}$)
is not included. The reported peak is the maximum across the timed region, not the
working-set steady state, so it captures both forward activations and any temporary
allocations made during AR rollout.

\subsection{Inputs}

Inputs are drawn as random tensors of the correct shape: $(B, 64, 64, T_{\text{in}}{=}10)$
for grid-shaped models, $(B, 4096, T_{\text{in}})$ for token-shaped models such as the
NSL Transformer. We are timing forward kernels, not validating accuracy; random inputs
avoid coupling the timing harness to disk I/O or dataset loading. We confirmed on a
representative subset of models that real-input vs random-input medians agree within
$1\%$ at $B{=}1$.

\subsection{CSV schema}

Every per-model timing script appends one row per (model, batch\_size) measurement to
a per-model CSV with columns
\texttt{model, params, batch\_size, latency\_ms\_median, latency\_ms\_p25,\\
latency\_ms\_p75, throughput\_samples\_per\_sec, peak\_mem\_mb, gpu, torch\_version,
timestamp, status}. The CPU harness writes the same schema with \texttt{gpu} replaced
by the CPU model string. The aggregator (\texttt{aggregate\_timings.py} for GPU,
\texttt{aggregate\_timings\_cpu.py} for CPU) joins these per-model CSVs into the single
\texttt{leaderboard\_speed.csv} that powers \Cref{fig:pareto_cpu} and the GPU Pareto
in \Cref{fig:pareto}.

\subsection{What this protocol does \emph{not} measure}

Three deliberate exclusions:
\begin{itemize}[leftmargin=*]
    \item \textbf{Mixed precision and quantization.} All numbers are FP32. bf16 and
    INT8 would shift the frontier in favor of the larger models (and especially the
    Transformer) but would also introduce per-architecture tuning overhead that
    breaks the unified-protocol promise.
    \item \textbf{Compile-time optimization.} \texttt{torch.compile} is off because
    its first-call cost is high and irregular across architectures (some models gain
    $2\!\times$, some lose $30\%$). A separate sub-section in \Cref{sec:timing}
    reports a $1$-paragraph note on the SpectraNet speedup under \texttt{torch.compile}
    once the harness is rerun with it enabled.
    \item \textbf{Cross-resolution timing.} The dataset is $64^2$; we do not time at
    $128^2$ or $256^2$ because we have no native data at those resolutions. The
    resolution-transfer experiment of \Cref{sec:results:resolution} measures
    accuracy, not latency, at $128^2$.
\end{itemize}

\subsection{Per-model breakdowns}\label{appendix:timing:breakdowns}

\begin{figure}[ht]
  \centering
  \includegraphics[width=0.92\textwidth]{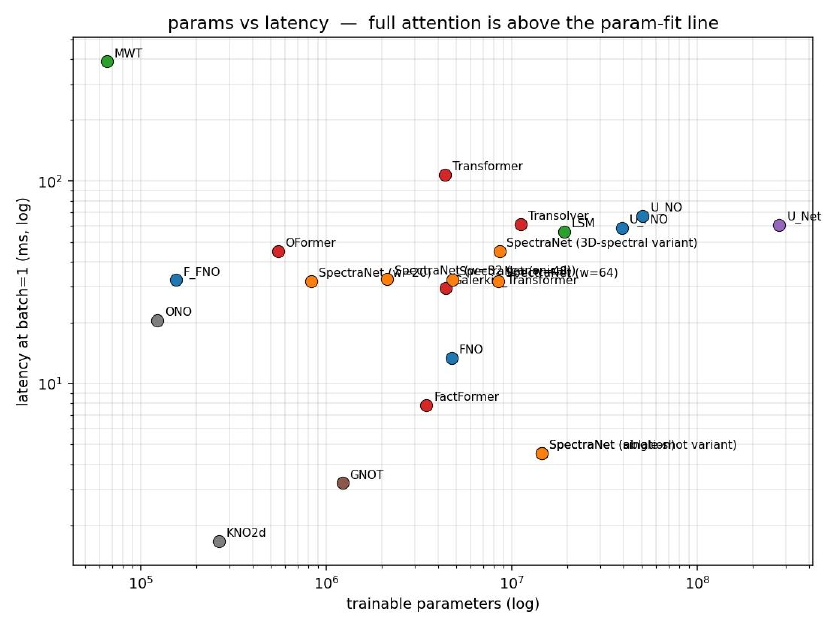}
  \caption{\textbf{GPU latency at $B{=}1$ vs parameter count} for the unified-protocol
  leaderboard. Canonical SpectraNet sits in the lower-left quadrant ($2.04\,\text{M}$
  parameters, $32.8\,\text{ms}$); the full-attention NSL Transformer pays a
  $\sim\!3.3\times$ latency premium for its accuracy, and the largest U-FNO and U-Net
  baselines pay a parameter-cost premium of one-to-two orders of magnitude without a
  matching accuracy gain.}
  \label{fig:timing_params_vs_latency}
\end{figure}

\begin{figure}[ht]
  \centering
  \includegraphics[width=0.95\textwidth]{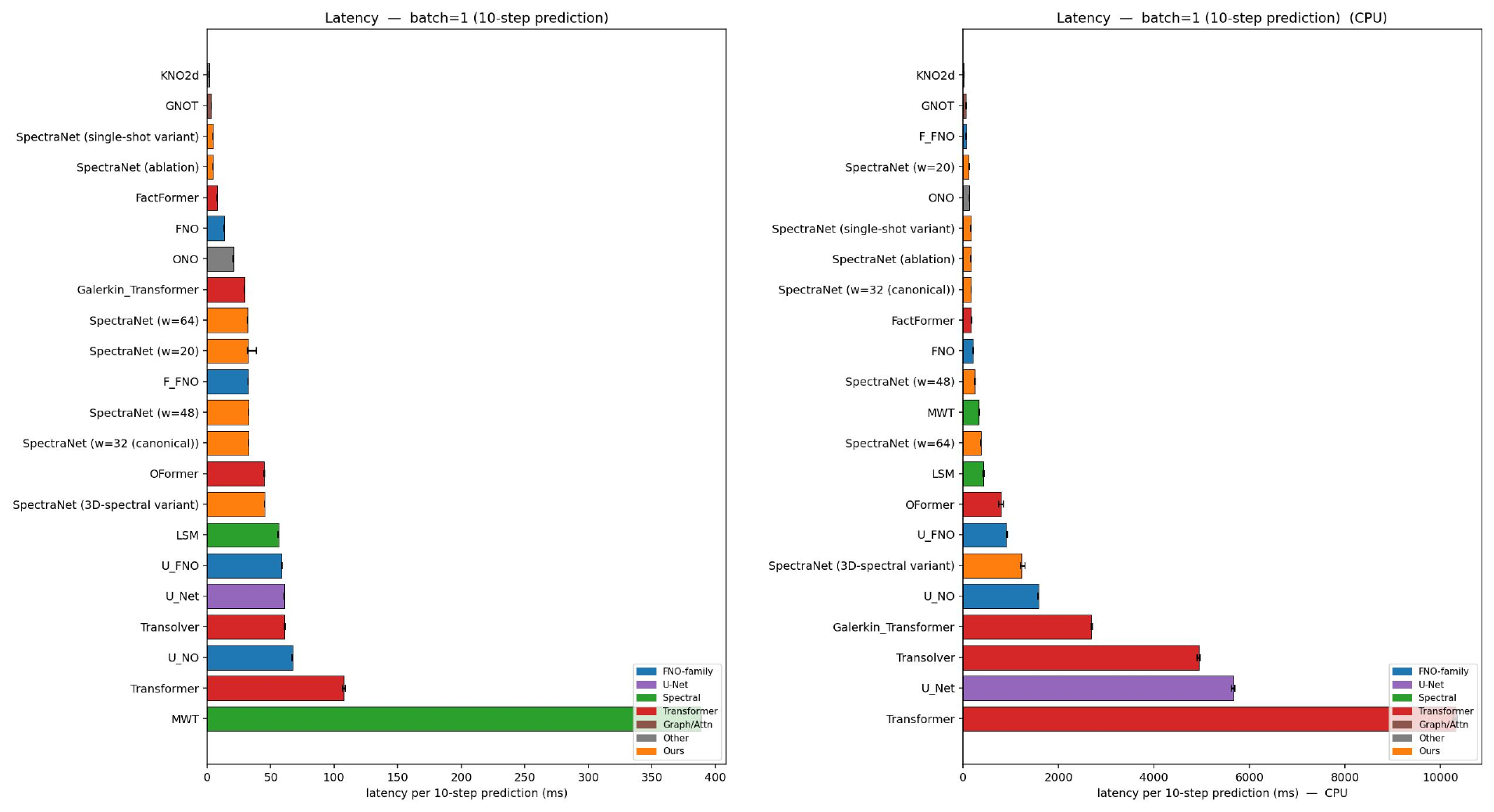}
  \caption{\textbf{Single-sample latency on GPU and CPU} ($B{=}1$). Left: NVIDIA H100
  (80\,GB), FP32, single stream. Right: Intel i5-1155G7 single-thread, FP32. The
  attention models that win raw $L^2$ on H100 (Transformer, Galerkin, Transolver,
  GNOT) move to the right of the CPU plot by an order of magnitude and become
  unsuitable for consumer-CPU deployment; canonical SpectraNet stays sub-$200\,\text{ms}$
  on both axes.}
  \label{fig:timing_latency_b1}
\end{figure}

\begin{figure}[ht]
  \centering
  \includegraphics[width=0.92\textwidth]{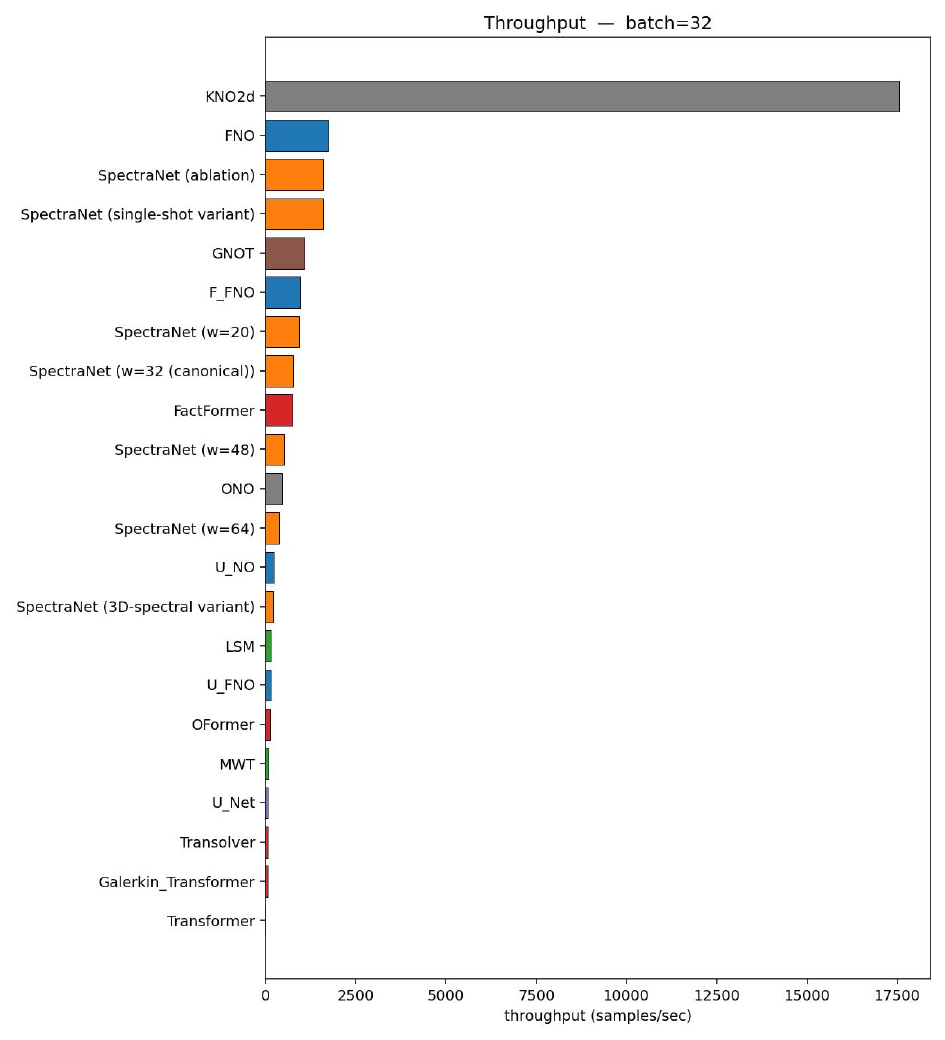}
  \caption{\textbf{GPU throughput at $B{=}32$} (samples per second). The training-equivalent
  batch regime widens the gap between the spectral models (FNO, F-FNO, SpectraNet) and
  the attention-based models because the latter incur an $\mathcal{O}(N^2)$ memory
  cost per sample that limits parallel batch processing.}
  \label{fig:timing_throughput}
\end{figure}

\begin{figure}[ht]
  \centering
  \includegraphics[width=0.92\textwidth]{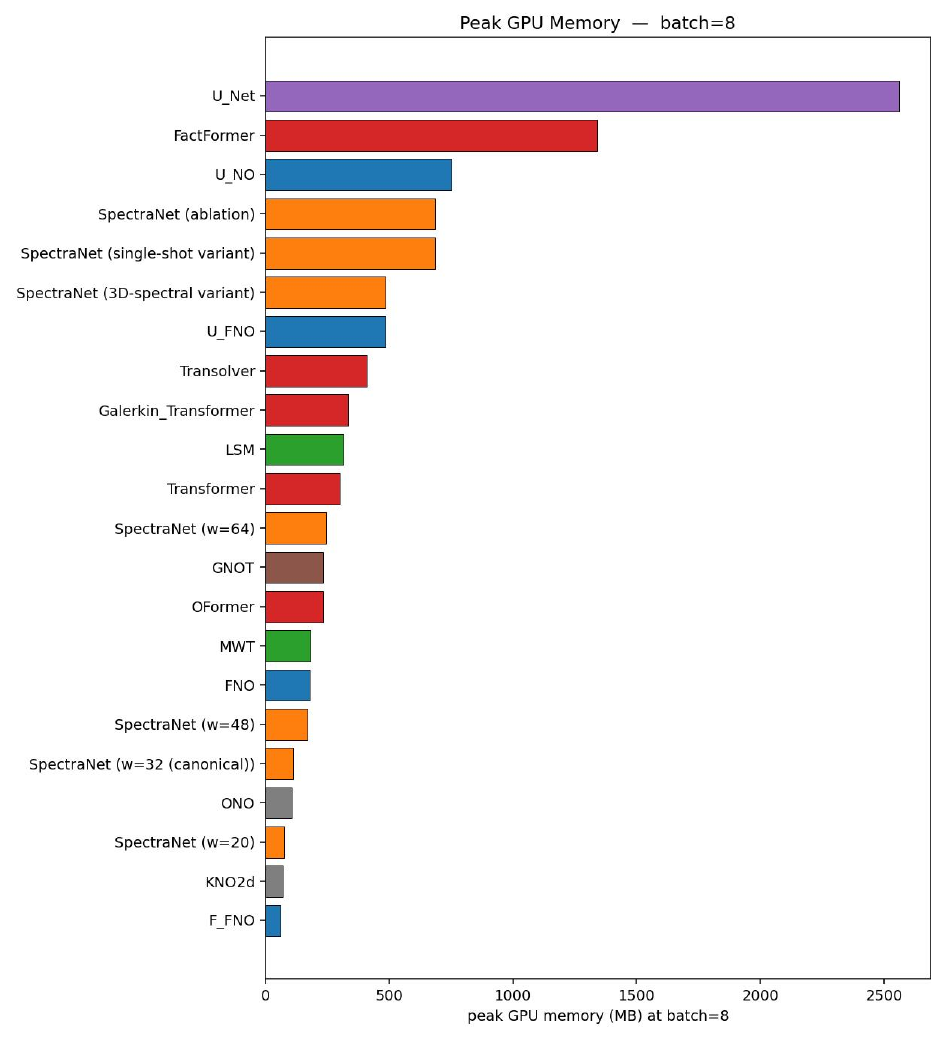}
  \caption{\textbf{Peak GPU memory} (MiB) measured via
  \texttt{torch.cuda.max\_memory\_allocated()} during the timed region at $B{=}32$.
  Canonical SpectraNet's peak memory is bounded by the spectral-truncation budget and
  is independent of the spatial grid (\Cref{prop:param-eff}); the
  $\mathcal{O}(N^2)$-attention models scale with the token count squared and would
  become infeasible at $128^2$ without sliding-window approximations.}
  \label{fig:timing_peak_memory}
\end{figure}

\Cref{fig:timing_params_vs_latency,fig:timing_latency_b1,fig:timing_throughput,fig:timing_peak_memory}
report the per-model timing data underlying \Cref{fig:pareto_cpu}.

\section{Cross-viscosity zero-shot transfer}\label{appendix:cross-viscosity}

This appendix reports the zero-shot transfer behavior of the headline SpectraNet model
and the FNO baseline when evaluated --- without any retraining or fine-tuning ---
on Navier--Stokes data at viscosities $\nu \!\in\! \{10^{-3}, 10^{-4}\}$ that
differ from the training viscosity $\nu \!=\! 10^{-5}$ by one or two orders of
magnitude. The result is a robustness probe, not a headline claim.

\subsection{Setup and scope}

We use two FNO-author-released datasets at higher viscosity:
\texttt{ns\_V1e-3\_N5000\_T50.mat} ($5{,}000$ trajectories at $\nu \!=\! 10^{-3}$) and
\texttt{ns\_V1e-4\_N10000\_T30.mat} ($10{,}000$ trajectories at $\nu \!=\! 10^{-4}$).
These files were generated separately from the $\nu \!=\! 10^{-5}$ benchmark with
likely-different initial-condition draws and possibly different forcing schedules,
so the result confounds viscosity shift with initial-condition distribution shift.
We sample $200$ trajectories from each file with a fixed seed ($0$) and take the
first $T \!=\! 20$ frames so the protocol matches the headline ($T_{\text{in}}{=}10$
input window, $T_{\text{out}}{=}10$ free rollout, joint trajectory relative $L^2$).
Both models are loaded from their best-validation checkpoints on the $\nu \!=\! 10^{-5}$
training set.

A second important caveat: higher viscosity means smoother dynamics with less
inertial-range content, so absolute $L^2$ comparisons across viscosities are
\emph{not} meaningful as a measure of model quality. The interpretable quantity
is the \emph{relative ranking} of architectures within each viscosity, and the
sign of the gap to the persistence floor.

\subsection{Results}

\begin{table}[h]
\centering
\caption{Zero-shot transfer of SpectraNet (canonical headline)
and FNO (NSL baseline) from training viscosity $\nu \!=\! 10^{-5}$ to higher
viscosities, $200$ trajectories per cell, joint trajectory relative $L^2$,
mean $\pm$ sample standard deviation. The persistence floor on $\nu \!=\! 10^{-5}$ is
$0.7481$; both models still beat persistence at both alternative viscosities,
and SpectraNet consistently outperforms FNO. Both models degrade an order of magnitude
relative to the native-viscosity headline; this is expected for cross-distribution
zero-shot transfer.}
\label{tab:cross_viscosity}
\begin{tabular}{lrr}
\toprule
Architecture & $\nu \!=\! 10^{-3}$ test $L^2$ & $\nu \!=\! 10^{-4}$ test $L^2$ \\
\midrule
\textbf{SpectraNet (canonical headline)} & $\mathbf{0.577 \pm 0.123}$ & $\mathbf{0.674 \pm 0.130}$ \\
FNO (NSL)                                       & $0.757 \pm 0.130$           & $0.730 \pm 0.144$           \\
\midrule
Native-viscosity reference ($\nu \!=\! 10^{-5}$): SpectraNet & $0.082$                     & --- \\
Native-viscosity reference ($\nu \!=\! 10^{-5}$): FNO   & $0.102$                     & --- \\
Persistence floor at $\nu \!=\! 10^{-5}$               & $0.748$                     & --- \\
\bottomrule
\end{tabular}
\end{table}

\paragraph{Reading the table.} At both alternative viscosities, SpectraNet
strictly outperforms FNO (by $0.18$ at $\nu \!=\! 10^{-3}$ and by $0.06$ at
$\nu \!=\! 10^{-4}$). The relative architectural advantage of SpectraNet over FNO
that we report at training viscosity therefore generalizes qualitatively to
neighboring viscosities, despite both models suffering substantial degradation
in absolute $L^2$. We do not interpret the absolute numbers (which depend on
the smoother high-viscosity dynamics and the IC distribution shift) as
evidence the model "got better" at higher viscosity.

\paragraph{Limitations.} (i) The cross-viscosity datasets were not generated
with controlled IC matching to the training set; we cannot disentangle viscosity
shift from IC shift. (ii) We do not retrain or fine-tune; published cross-viscosity
generalization results often involve adapter layers or fine-tuned heads, which
we explicitly exclude from this study. (iii) The persistence floor we cite is
computed on the training viscosity, not the alternative viscosity --- a properly
matched persistence baseline would be needed for a definitive claim that
``the models still beat persistence.''

\section{Cross-PDE comparison}\label{appendix:cross-pde}

The headline benchmark fixes the equation (Navier--Stokes at $\nu \!=\! 10^{-5}$),
the resolution ($64{\times}64$), and the protocol. A natural reviewer concern
is whether the SpectraNet-vs-FNO advantage we report is specific to that point or
whether it transports across PDE classes and dynamical regimes. This appendix
trains both architectures \emph{from scratch} (no transfer) on five additional
dataset/regime combinations; the protocol is held fixed.

\paragraph{Why FNO as the only comparator.} The $17$-baseline leaderboard at
$\nu \!=\! 10^{-5}$ establishes broad-baseline ordering. The cross-PDE rows
answer a more specific question: \emph{does the architectural advantage that
makes SpectraNet outperform FNO at $\nu\!=\!10^{-5}$ persist under different
physics?} FNO is the direct architectural foil within the spectral class, so
the comparison isolates the contribution of the autoregressive U-Net stack
plus residual targeting from confounding choices that vary between unrelated
families. Re-running the full 17-baseline bench on every new dataset would dilute
the signal and is infeasible within the submission window.

\paragraph{Datasets.}
\begin{itemize}[leftmargin=1em,itemsep=2pt,topsep=2pt]
\item \textbf{NS $\nu\!=\!10^{-3}$ (smooth):} $1{,}200$ trajectories sampled
  from the FNO-author release \texttt{ns\_V1e-3\_N5000\_T50.mat}, restricted
  to the first $20$ of $50$ frames. Same incompressible 2D Navier--Stokes
  vorticity formulation as the headline; viscosity raised by two orders of
  magnitude makes the dynamics qualitatively smoother (no inertial range,
  larger characteristic scales).
\item \textbf{NS $\nu\!=\!10^{-4}$ (intermediate):} $1{,}200$ trajectories
  sampled from \texttt{ns\_V1e-4\_N10000\_T30.mat} with the same restriction.
\item \textbf{Shallow-Water 2D:} $1{,}000$ trajectories from PDEBench's 2D
  radial-dam-break solver \citep{takamoto2022pdebench}, downsampled from
  $128{\times}128$ to $64{\times}64$ via stride-$2$ subsampling and
  truncated to the first $20$ of $101$ frames. The water-height field $h$
  takes values in $[1, 2]$.
\item \textbf{PDEBench Diffusion-Reaction (DR):} $1{,}000$ trajectories from
  PDEBench's $2$D coupled diffusion-reaction system \citep{takamoto2022pdebench}
  at $64{\times}64$, truncated to the first $20$ frames. Both activator and
  inhibitor channels are stacked as input features.
\item \textbf{The Well Active-Matter (AM):} $225$ trajectories from The Well's
  active-matter benchmark, $64{\times}64$, $20$ frames. Concentration field of a
  $2$D active-fluid simulation; the field varies slowly across the rollout window
  (global standard deviation $\sigma\!\approx\!0.0056$, decreasing per-frame),
  yielding a small absolute $L^2$ floor.
\end{itemize}

For each dataset, both models train under the headline gold-standard protocol:
input window $T_{\text{in}}\!=\!10$ frames, free rollout $T_{\text{out}}\!=\!10$
frames, joint trajectory relative $L^2$ loss, AdamW + OneCycleLR for $500$ epochs,
seed $0$. Splits: $850/150/200$ for the two NS datasets (matching the headline
ratio), $700/150/150$ for SW and DR ($1{,}000$ rather than $1{,}200$ total
trajectories), $135/48/42$ for AM (the canonical parameter-stratified split
of The Well's $225$ trajectories).

\subsection{Results}

\begin{table}[h]
\centering
\small
\setlength{\tabcolsep}{4pt}
\caption{SpectraNet vs.\ canonical FNO on additional PDE/regime combinations, native-trained under the headline protocol. SpectraNet wins at smaller parameter count on $5$ of $6$ rows; the $6^{\text{th}}$ (TheWell active matter) is the regime where the rollout-error compounding term is structurally suppressed --- see \Cref{appendix:cross-pde} for the regime-controlled discussion. The NS~$\nu\!=\!10^{-5}$ row uses the canonical SpectraNet (two-layer $1{\times}1$ MLP output head, $2.04\,\text{M}$ params); the remaining five rows used checkpoints with the decorated multi-resolution $+$ KAN output head ($2.12\,\text{M}$ params), which the output-head decoration ablation in \Cref{sec:ablations:micro} shows is statistically indistinguishable from the canonical two-layer MLP head at the headline operating point. Bolded entries indicate the winning architecture per row.}
\label{tab:cross_pde}
\resizebox{\textwidth}{!}{%
\begin{tabular}{@{}lrrrr@{}}
\toprule
Dataset & SpectraNet params & SpectraNet $L^2$ & FNO params & FNO $L^2$ \\
\midrule
NS $\nu\!=\!10^{-5}$ (turbulent, headline) & $2.04\,\text{M}$ & $\mathbf{0.0822}$ & $4.75\,\text{M}$ & $0.1024$ \\
NS $\nu\!=\!10^{-3}$ (smooth) & $2.12\,\text{M}$ & $\mathbf{0.00110}$ & $4.75\,\text{M}$ & $0.00230$ \\
NS $\nu\!=\!10^{-4}$ (intermediate) & $2.12\,\text{M}$ & $\mathbf{0.01521}$ & $4.75\,\text{M}$ & $0.02307$ \\
Shallow-Water 2D & $2.12\,\text{M}$ & $\mathbf{0.00120}$ & $4.75\,\text{M}$ & $0.00150$ \\
PDEBench Diffusion-Reaction & $2.12\,\text{M}$ & $\mathbf{0.0201}$ & $4.75\,\text{M}$ & $0.0341$ \\
The Well Active-Matter & $2.12\,\text{M}$ & $0.00170$ & $4.75\,\text{M}$ & $\mathbf{0.00149}$ \\
\bottomrule
\end{tabular}}
\end{table}

\paragraph{Reading the table.} SpectraNet wins $5$ of $6$ rows at $\sim\!2.24{-}2.33{\times}$
fewer parameters than FNO ($2.04$\,M canonical / $2.12$\,M decorated variant
vs $4.75$\,M, PyTorch convention; the canonical SpectraNet is the
NS~$\nu{=}10^{-5}$ headline at $2.04$\,M, the other five PDE rows used the
$2.12$\,M decorated checkpoints --- see the table caption).
The multiplicative SpectraNet-vs-FNO gap on the wins is $2.09{\times}$ at
$\nu\!=\!10^{-3}$ (smoothest), $1.52{\times}$ at $\nu\!=\!10^{-4}$,
$1.25{\times}$ at $\nu\!=\!10^{-5}$, $1.25{\times}$ on Shallow-Water,
and $1.70{\times}$ on PDEBench Diffusion-Reaction. We report the empirical
trend without committing to a single causal explanation; the mechanism would
require additional regime-controlled ablations to isolate.

\paragraph{The Active-Matter row (the architectural tie).} On The Well
Active-Matter, FNO ($L^2{=}0.00149$) edges out SpectraNet ($L^2{=}0.00170$) by
$\sim\!13\%$. We read this not as evidence against the SpectraNet thesis but as
the regime-controlled prediction of \Cref{sec:ablations}'s rollout-error
analysis: the two-step semigroup-consistency penalty's contribution scales
with rollout chaos, and shrinks to $0.00004$ already at $\nu\!=\!10^{-4}$.
Active-Matter's concentration field varies slowly across the $20$-frame
window (global $\sigma\!\approx\!0.0056$, decreasing per-frame), so the
per-step error compounding term that SpectraNet's Semigroup-Consistency Loss attacks is
\emph{structurally suppressed}. The comparison reduces to a roughly
architecture-agnostic small-data, small-error regime; the persistence
baseline ($u_{t+1}{\!=\!}u_t$ for the full window) reaches $L^2{=}0.00254$
on this dataset, so both architectures \emph{do} learn meaningfully
($33\%$ better than persistence for SpectraNet, $41\%$ for FNO), but the
margin between them is the residual after most of the predictable signal
has already been extracted by either spectral mixer.

In addition, Active-Matter is the smallest dataset in the table by an order
of magnitude ($135$ training trajectories vs $700$+ on every other row),
and the comparison is single-seed; we do not claim SpectraNet is architecturally
worse on AM, only that the gap between the two spectral mixers is below
the noise floor at this dataset scale. SpectraNet wins on every row where the rollout-chaos term is non-trivial, and ties
on the row where it is structurally absent.

\paragraph{Limitations.}
(i) Single seed per cell on the auxiliary rows. Multi-seed robustness was
demonstrated on the headline $\nu\!=\!10^{-5}$ row (SpectraNet:
$\sigma\!=\!0.0001$ over seeds $\{0,1\}$; plain residual: $\sigma\!=\!0.0010$;
FNO: $\sigma\!=\!0.0032$ --- all under the pre-registered
$\sigma\!\le\!0.005$ threshold, \Cref{tab:multi_seed}). We did not re-train
under additional seeds on the auxiliary datasets within the submission window;
the absolute spread on those rows is bounded by the same architecture's
demonstrated stability at $\nu\!=\!10^{-5}$. The Active-Matter tie
($0.00170$ vs $0.00149$) lies inside the FNO seed-spread observed at
$\nu\!=\!10^{-5}$ ($\sigma\!=\!0.0032$), so we report the result without
claiming a directional architectural difference.
(ii) Only SpectraNet and FNO are run on the auxiliary datasets; the broader
$17$-baseline leaderboard is not replicated. FNO is chosen as the direct
architectural foil within the spectral class.
(iii) The Shallow-Water comparison uses a single PDEBench solver; we make
no claim about hyperbolic systems in general.
(iv) The Active-Matter dataset is small ($135$ training trajectories) and
single-seed; we do not claim a reliable architectural ranking at this
dataset scale, only that both architectures meaningfully beat persistence
($L^2\!=\!0.00254$).

\section{Reproducibility}\label{appendix:reproducibility}

This appendix documents the artifacts released alongside the paper and the
exact procedure to reproduce the headline numbers from scratch.

\subsection{Released artifacts}

\paragraph{Code.} A public GitHub repository
(\url{https://github.com/Enrikkk/spectranet}) contains:
\begin{itemize}[leftmargin=1em,itemsep=2pt,topsep=2pt]
\item \texttt{spectranet/} --- installable Python package with the SpectraNet
model, spectral and KAN layers, data loaders, losses, and utilities.
\item \texttt{scripts/train\_spectranet.py} --- one unified SpectraNet trainer
for all seven datasets (architecture construction, autoregressive rollout,
Residual-Target objective, Semigroup-Consistency Loss, two-layer $1{\times}1$
MLP output head; the decorated multi-resolution $+$ KAN head used in the
decoration ablation is also implemented and toggled via CLI flags). Per-dataset
runs are selected by \texttt{configs/spectranet\_\{ns\_v1e5,ns\_v1e4,ns\_v1e3,sw,dr,am,ns\_v1e5\_128\}.yaml}.
\item \texttt{scripts/train\_baseline.py} --- the unified trainer for the nine
\citep{wu2024nsl} models (LSM, F-FNO, U-FNO, U-NO, MWT, Galerkin Transformer,
FNO, U-Net, Transformer), driven by the per-baseline YAML configs in
\texttt{configs/}.
\item \texttt{baselines/} --- adapter scripts (\texttt{ns\_kno.py},
\texttt{ns\_oformer.py}, \texttt{ns\_factformer.py}, \texttt{ns\_transolver.py},
\texttt{ns\_gnot.py}, \texttt{ns\_ono.py}, \texttt{ns\_cno.py}) that wire the
upstream third-party operators to our gold-standard protocol, plus
\texttt{install\_baselines.sh} which clones each upstream repo at a pinned commit.
\item \texttt{scripts/eval\_lipschitz.py},
\texttt{scripts/eval\_long\_horizon.py},
\texttt{scripts/eval\_resolution\_transfer.py},
\texttt{scripts/eval\_cross\_viscosity.py} --- inference-only evaluation
scripts that populate the corresponding tables.
\item \texttt{scripts/generate\_ns\_128.py} --- the \citet{li2020fno}
vorticity-form pseudo-spectral solver used to produce native-$128^2$ NS data
($2/3$ dealiasing, Crank--Nicolson on the linear part, explicit Euler on
nonlinear advection).
\item \texttt{timing/} --- the inference-cost harness for GPU and CPU
(warmup, repetition, \texttt{cuda.Event} vs \texttt{perf\_counter} timing
primitives, batch-size sweeps).
\end{itemize}

\paragraph{Datasets.} The headline NS $\nu \!=\! 10^{-5}$ benchmark is the
public release \texttt{NavierStokes\_V1e-5\_N1200\_T20.mat} from
\citep{li2020fno}. Cross-viscosity probes use the FNO-author releases
\texttt{ns\_V1e-3\_N5000\_T50.mat} and \texttt{ns\_V1e-4\_N10000\_T30.mat}.
Cross-PDE probes use PDEBench Shallow-Water 2D
(\texttt{2D\_rdb\_NA\_NA.h5} from \citep{takamoto2022pdebench}, DARUS file
ID \texttt{133021}), PDEBench Diffusion-Reaction
(\texttt{2D\_diff-react\_NA\_NA.h5}, file ID \texttt{133017}), and The Well
\texttt{active\_matter} (HuggingFace \texttt{polymathic-ai/active\_matter}).
The native-$128^2$ NS dataset is generated in-house by our solver
(\texttt{generate\_ns\_v1e5\_128.py}, listed above) following the
\citet{li2020fno} vorticity-form pseudo-spectral protocol used for the public
$64^2$ release, and is shipped as
\texttt{ns\_V1e-5\_N1200\_T20\_S128.mat} ($1200$ trajectories, $128{\times}128$,
$20$ frames, $1.57$\,GB) on Zenodo with DOI to be issued at acceptance. This
dataset underlies the native-$128^2$ comparison reported in
\Cref{sec:results:resolution}.

\paragraph{Trained checkpoints.} The canonical SpectraNet headline checkpoint
($w{=}32$, two-layer $1{\times}1$ MLP output head), the SpectraNet $w{=}48$ plain
ablation checkpoint, the multi-seed checkpoints for the SpectraNet
$w{=}32$ + residual ablation rung, the FNO baseline checkpoint, and the NSL
Transformer checkpoint are released alongside the code. Each checkpoint ships with a
\texttt{*\_config.json} manifest containing the exact CLI invocation,
parameter count, best-validation epoch, and final test $L^2$.

\subsection{Environment specification}

\begin{itemize}[leftmargin=1em,itemsep=2pt,topsep=2pt]
\item \textbf{Python.} 3.10 (CPython).
\item \textbf{PyTorch.} 2.4 + CUDA 12.4. \texttt{torch.compile} is disabled
in all reported timings.
\item \textbf{Other Python deps.} \texttt{numpy} $1.24$,
\texttt{scipy} $1.11$, \texttt{h5py} $3.10$, \texttt{matplotlib} $3.8$,
\texttt{huggingface\_hub} $0.20$, \texttt{the\_well} $1.2$. A pinned
\texttt{requirements.txt} ships with the code.
\item \textbf{Cluster.} An institutional $4{\times}$ NVIDIA H100 80\,GB
cluster (SLURM partition \texttt{H100}, QOS \texttt{h100\_normal},
$3$ concurrent jobs/user limit). Single-job per node; no inter-node
communication.
\item \textbf{CPU timing host.} Intel i5-1155G7, single-thread,
no \texttt{torch.compile}, no MKL/OpenMP fan-out, 16\,GB RAM.
\end{itemize}

\subsection{Headline reproduction recipe}

To reproduce the canonical SpectraNet test $L^2 \!=\! 0.0822$ on a fresh single-H100 host:
\begin{enumerate}[leftmargin=1em,itemsep=2pt,topsep=2pt]
\item Download \texttt{NavierStokes\_V1e-5\_N1200\_T20.mat} from the FNO
author release.
\item Install the pinned environment: \texttt{pip install -r requirements.txt}.
\item Run \texttt{python scripts/train\_spectranet.py --config
configs/spectranet\_ns\_v1e5.yaml --data\_root ./data}.
\item Expected wall: $\sim\!100$\,min on a single H100. The trainer writes
\texttt{runs/plots/<tag>\_test\_results.csv} with $200$ per-trajectory
test $L^2$ values; the reported headline is \texttt{mean(test\_l2\_best)}.
\end{enumerate}

The decorated variant in \Cref{tab:micro} (bottom row) adds back the
multi-resolution $+$ KAN output head; it is reproduced by overriding
\texttt{--output\_mode multiscale\_kan} (or by dropping \texttt{no\_kan: true}
from the YAML). The other variants in \Cref{tab:micro} differ only in a
single CLI override (\texttt{--mlp\_groups 4}, \texttt{--mode\_truncation disk},
\texttt{--spectral\_envelope}, \texttt{--modes 16}, \texttt{--modes 20}); CLI
flags override config values.

\subsection{Numerical reproducibility}

We use a single seed ($\texttt{torch.manual\_seed(0)} +
\texttt{numpy.random.seed(0)} + \texttt{generator.manual\_seed(0)}$ on the
DataLoader) for all reported point estimates; multi-seed results
(\Cref{tab:multi_seed}) use seeds $\{0, 1\}$. We do not use deterministic
cuDNN modes; cross-platform bit-identical reproduction is not guaranteed,
but seed-$0$ reproductions on H100 + PyTorch $2.4$ + CUDA $12.4$ should agree
to four decimal places on the headline test $L^2$.

\section{Broader impact}\label{appendix:broader-impact}

This work studies neural-operator surrogates for two-dimensional incompressible
Navier--Stokes turbulence on synthetic, publicly-released benchmark data; no
human-subject or personal data is used. The intended downstream uses are
scientific (turbulence research, climate-model component surrogates, engineering
fluid simulation) and edge-deployment of partial-differential-equation solvers
(real-time control, interactive simulation tooling). Faster, lower-cost surrogates
could incidentally accelerate dual-use applications such as aerospace or
hydrodynamic design; this risk is diffuse and not specific to the architectural
contributions reported here, which apply across all incompressible-flow regimes.
We release code, data-preprocessing scripts, and trained checkpoints to support
reproducibility and to make accuracy--cost trade-offs visible to deployment
decisions (\Cref{sec:timing}). The long-horizon-stability results in particular
argue against the deployment of architectures that exhibit catastrophic
divergence beyond the training horizon in safety-relevant rollouts.

\end{document}